\documentclass[conference]{IEEEtran}
\IEEEoverridecommandlockouts
\usepackage{cite}
\usepackage{amsmath,amssymb,amsfonts}
\usepackage{algorithmic}
\usepackage{graphicx}
\usepackage{textcomp}
\usepackage{xcolor}
\usepackage{mathrsfs}
\usepackage{adjustbox}

\usepackage{algorithm}
\usepackage{algorithmic}
\usepackage{amsthm}
\usepackage{siunitx}
\usepackage{subcaption}
\usepackage{booktabs}
\usepackage{listings}
\usepackage{mathtools}
\usepackage{tikz}
\usetikzlibrary{bayesnet}
\usepackage{tkz-tab}
\usepackage{witharrows}
\usepackage{stmaryrd}
\usepackage{url}
\theoremstyle{plain}
\newtheorem{theorem}{Theorem}[section]

\newtheorem{lemma}[theorem]{Lemma}

\theoremstyle{definition}

\theoremstyle{remark}

\def\BibTeX{{\rm B\kern-.05em{\sc i\kern-.025em b}\kern-.08em
    T\kern-.1667em\lower.7ex\hbox{E}\kern-.125emX}}
\begin{document}
% Authors macros
\def\cmjo#1{\textcolor{cyan}{#1}}
\def\cjul#1{\textcolor{olive}{#1}}
\def\ctim#1{\textcolor{orange}{#1}}
\def\rqe#1{\textcolor{red}{#1}}
\def\mjo#1{\todo[inline, color=cyan, size=normalsize]{@mjo: #1}}
\def\jul#1{\todo[inline, color=olive, size=normalsize]{@jul: #1}}
\def\tim#1{\todo[inline, , size=normalsize]{@tim: #1}}
\def\seb#1{\todo[inline, color=magenta, , size=normalsize]{@seb: #1}}
\def\ulr#1{\todo[inline, color=lime, , size=normalsize]{@ulr: #1}}

% ML and Rule Lists (background)
\def\dataset{\mathcal{D}}
\def\featset{\mathcal{F}}
\def\nsamples{n}
\def\nfeatures{m}

\def\asample{s}
\def\asampleind#1{s_{#1}}
\def\afeatind#1{x_{#1}}
\def\alabel{y}
\def\alabelind#1{y_{#1}}

\def\cart{\texttt{CART}}

\newcommand{\ruleset}{\mathscr{R}} %ruleset (pre-mined)
\def\rlist{RL} %rule list
\def\arule{r} %rule
\def\aruleind#1{r_{#1}} %rule_i

% DP (background)
\newcommand{\universe}{\mathscr{X}}
\newcommand{\databases}{\mathbb{N}^{|\universe|}} %database
\newcommand{\algo}{\mathcal{M}} %algo
\newcommand{\mechan}{\mathscr{M}} %DP mechanism ?
\def\amechanind#1{m_{#1}} %rule_i
\def\noise{N}
%function
%utility
\def\utilrange{\mathcal{V}}
% r ... range

% Greedy RL and Gini impurity
\def\GRL{\texttt{GreedyRL}}
\def\setcapt#1{C(#1)} % subset of samples: captured
\def\ncapt#1{n_c(#1)} % nb samples: captured by a rule
\def\nleft#1{n_l(#1)} % nb samples: left by a rule
\def\nremain#1{\tilde{\nsamples}(#1)} %nn samples left to be captured
\def\labelpred#1#2{\hat{y}_{#2}(#1)}

\def\gini#1{\mathscr{G}(#1)} %Gini impurity
\def\ginicapt#1{\mathscr{G}_c(#1)}
\def\ginileft#1{\mathscr{G}_l(#1)}

\def\minsupp{\Lambda}

% Smooth Sensitivity
\def\ls{LS} %local sensitivity
\def\smoothub{S}
\def\smooth{S^*}

\def\DPGRL{\texttt{sm-Laplace}}

\newcommand{\norm}[2]{\left\lVert#2\right\rVert_{#1}}
\newcommand{\equivasym}[1]{\underset{#1}{\sim}}
\newcommand{\floor}[1]{\lfloor #1 \rfloor}
\newcommand{\ceil}[1]{\lceil #1 \rceil}

\newcommand{\namedataset}[1]{\texttt{#1}}

\newcommand{\p}{\mathbb{P}} %probability
\newcommand{\ind}{\mathbb{1}} %indicatrice
\newcommand{\capt}{\text{cap}} %elts capturés
\newcommand{\supp}{\text{supp}} %support 

\newcommand{\argmax}{\mathop{\mathrm{argmax}}\limits}  

% Rule lists display
\lstdefinelanguage{RuleListsLanguage}{
  keywords={if, then, else},
  keywordstyle=\color{blue}\bfseries,
  ndkeywords={True, False},
  ndkeywordstyle=\color{red}\bfseries,
  identifierstyle=\color{black},
  sensitive=false,
  comment=[l]{//},
  morecomment=[s]{/*}{*/},
  commentstyle=\color{purple}\ttfamily,
  stringstyle=\color{red}\ttfamily,
  morestring=[b]',
  morestring=[b]"
}

\definecolor{RuleListsLanguageBackgroundColor}{rgb}{0.95, 0.95, 0.95}

%ajouter fct pour algo

\title{Smooth Sensitivity for Learning Differentially-Private yet Accurate Rule Lists}

%\author{\IEEEauthorblockN{Anonymous Author(s)}\vspace{5pt}}
\author{
\IEEEauthorblockN{1\textsuperscript{st} Timothée Ly}
\IEEEauthorblockA{
\textit{LAAS-CNRS, Telecom Paris \& KTH}\\
Toulouse, France \\
tly@laas.fr}
\and
\IEEEauthorblockN{2\textsuperscript{nd} Julien Ferry}
\IEEEauthorblockA{
\textit{Polytechnique Montréal}\\
Montréal, Canada \\
julien.ferry@polymtl.ca}
\and
\IEEEauthorblockN{3\textsuperscript{rd} Marie-Jos\'e Huguet}
\IEEEauthorblockA{
\textit{LAAS-CNRS, Universit\'{e} de Toulouse, CNRS, INSA}\\
Toulouse, France \\
huguet@laas.fr}
\and
\IEEEauthorblockN{4\textsuperscript{th} S\'ebastien Gambs}
\IEEEauthorblockA{\textit{Universit\'e du Qu\'ebec \`a Montr\'eal} \\
Montr\'eal, Canada \\
gambs.sebastien@uqam.ca}
\and
\IEEEauthorblockN{5\textsuperscript{th} Ulrich A{\"i}vodji}
\IEEEauthorblockA{\textit{\'Ecole de Technologie Sup\'erieure} \\
Montr\'eal, Canada \\
Ulrich.Aivodji@etsmtl.ca}
}

\maketitle

\begin{abstract}
Differentially-private (DP) mechanisms can be embedded into the design of a machine learning algorithm to protect the resulting model against privacy leakage. However, this often comes with a significant loss of accuracy due to the noise added to enforce DP. % \\
In this paper, we aim at improving this trade-off for a popular class of machine learning algorithms leveraging the Gini impurity as an information gain criterion to greedily build interpretable models such as decision trees or rule lists. To this end, we establish the smooth sensitivity of the Gini impurity, which can be used to obtain thorough DP guarantees while adding noise scaled with tighter magnitude. % \\
We illustrate the applicability of this mechanism by integrating it within a greedy algorithm producing rule list models, motivated by the fact that such models remain understudied in the DP literature.
Our theoretical analysis and experimental results confirm that the DP rule lists models integrating smooth sensitivity have higher accuracy that those using other DP frameworks based on global sensitivity, for identical privacy budgets.
\end{abstract}


\section{Introduction}
\label{sec:intro}
Machine learning models are increasingly used for high-stakes decision-making tasks such as kidney exchange~\cite{DBLP:conf/aaai/0001CDM21} or recidivism prediction~\cite{angwin2016machine}. 
As such tasks often require the use of sensitive data (\emph{e.g.}, medical or criminal records), it is crucial to ensure that the learned models do not leak undesired information.
Another important aspect is to make sure human users can verify and trust the models' decisions, which motivates the use of inherently interpretable models when possible~\cite{rudin2019stop}. However, such models are also vulnerable to privacy attacks such as membership inference~\cite{membership-inference,carlini_mia}, in which the objective of the adversary is to infer the presence of a particular profile in the training dataset, or reconstruction attacks~\cite{ferry:hal-04189566,pmlr-v235-ferry24a}, in which the aim of the adversary is to reconstruct the training set. 
To counter this issue, Differential Privacy (DP)~\cite{DBLP:conf/tcc/DworkMNS06,foundations_DP} has emerged as a \emph{de facto} privacy standard, thoroughly bounding the amount of information any adversary can gain regarding any single individual in the dataset.
%Seb: j'ai mis en commentaires pour gagner de la place
%Seb: merci Timothée pour le travail je vais essayer de raccourcir sans toucher au coeur de l'article
%More precisely, DP aims at reconciling two antagonist purposes in privacy-preserving machine learning: extracting useful correlations from data without revealing private information about a particular individual. 

For instance, recent works~\cite{DBLP:journals/cim/GongXPFQ20,survey-ji} survey the literature on existing DP variants of different typical machine learning algorithms, such as the DP versions of the Principal Component Analysis algorithm~\cite{DPPCA} and of the Stochastic Gradient Descent~\cite{DPSGD}. 
However, much less work has been dedicated to the DP learning of interpretable models. Nonetheless, a line of works have studied different adaptations of learning algorithms producing tree-based models (\emph{i.e.}, mostly decision trees and random forests)~\cite{decision-tree}. These works rely on popular greedy induction algorithms such as \texttt{CART}~\cite{cart}. An important challenge they had to face is that the Gini impurity, a popular information gain criterion widely used at each iteration of these greedy algorithms, has very high global sensitivity. This leads to the addition of a considerable amount of noise to comply with DP, hence harming the utility of the resulting model.
In this paper, we address this issue by leveraging the notion of smooth sensitivity~\cite{nissim}.
%studying one important direction pointed out by
%establishing the smooth sensitivity of the Gini impurity. 
More precisely, we first theoretically characterize the smooth sensitivity of the Gini impurity. 
Then, we design a DP mechanism based on smooth sensitivity that we integrate into a greedy algorithm for learning rule list models. Depending on the considered noise distribution, our approach can provide either pure or approximate DP guarantees.
Our experimental results show that the proposed DP mechanisms incur a lower accuracy loss than other mechanisms based on global sensitivity for identical privacy budgets. 

The outline of the paper is as follows. 
First, in Section~\ref{sec:background}, we recall the background on rule lists models and DP. 
Afterwards, in Section~\ref{sec:art}, we introduce the building blocks of our approach, namely greedy learning of rule lists, Gini impurity and smooth sensitivity. % that form the basic of our solution.
Then, in Section~\ref{sec:contrib}, we present our main contribution on establishing the smooth sensitivity of the Gini impurity. We leverage it to design an effective algorithm for learning DP rule lists. 
Finally, in Section~\ref{sec:expe}, we empirically evaluate our proposed methods in terms of privacy-accuracy trade-offs and robustness to privacy attacks before concluding in Section~\ref{sec:concl}. 

%Ulr: j'ai commenté outline pour ganger de l'espace. on pourra le remettre sur la camera ready avec la page additionnelle
\section{Background}
\label{sec:background}
%In this section, we introduce rule lists models before presenting the background on differential privacy.
\subsection{Rule Lists} %\& Notations}
\label{subsec:background-rl}
We consider a tabular dataset $\dataset$ of $\nsamples$ samples in which each sample $\asample$ corresponds to a set of binary features and has 
a binary label $\alabelind{\asample}$. 
Rule lists, originally introduced as a way to efficiently represent Boolean functions, are a common type of interpretable models~\cite{Rivest, corels_1}.
More precisely, a rule list $\rlist$ is a sequence of $K+1$ rules $(\aruleind{1}, \dots, \aruleind{K}, \aruleind{0}) \in \ruleset^{K+1}$ in which $\ruleset$ is the set of possible rules (which, for instance, can be pre-mined). 
Each rule $\aruleind{i} \in \ruleset$ is composed of a Boolean assertion $p_i$ called the antecedent and a label prediction $q_i \in \{0,1\}$ named the consequent and can be denoted by $\aruleind{i} = p_i \rightarrow q_i$.  
A sample $\asample$ of dataset $\dataset$ is said to be caught by rule $\aruleind{i}$ when $p_{i}$ evaluates to true for $\asample$, which leads to $\asample$ being classified with label $q_i$. 
The \textit{default rule}, $\aruleind{0} = \text{True} \rightarrow q_0$ %is called the \textit{default rule} and 
classifies any sample not caught by the previous rules to $q_0 \in \{0,1\}$ fixed.
%Remark that the antecedent $p_i$ can be the logical conjunction of several Boolean assertions.
%\jul{A quoi sert la phrase ci-dessus ?}
%\tim{pour indiquer que la cardinalité dans les rules lists peut être > 1}
%\jul{Je veux bien simplifier ces notations et ré-écrire cette sous-section (en particulier tout ce qui n'est pas utilisé par la suite devrait disparaître (eg (je pense) $x_i$)) mais il faut qu'on en discute avec @Tim.}
%\seb{je suis d'accord avec la simplification}

%\seb{il manque la référence et ci-dessous}
%\jul{Si c'est de l'avantage en terme d'expressivité (à taille comparable) c'est le papier de rivest qu'il faut citer. Pour le côté cognition on en avait parlé mais la comparison est plutôt désavantageuse pour les RLs donc à voir si on dit quelque chose (ou pas ?) (cf. paragraphe bas de page 33 de ma thèse)}
Rule List~\ref{fig:rule-list} provides an example model trained for a recidivism prediction task.
As evidenced through this illustrative example, the use of any discriminatory feature would immediately be spotted, which is an advantage of using inherently interpretable models~\cite{rudin2019stop}. This is in contrast with black-box models, in which such an undesired behaviour would be difficult - or even impossible~\cite{DBLP:journals/corr/abs-1910-01432} - to detect. 

\lstset{numberbychapter=false,caption=\lstname,frame=single, stepnumber=1, numbersep=2pt, xleftmargin=0.01\linewidth, xrightmargin=0.01\linewidth}
\renewcommand{\lstlistingname}{Rule list}
\begin{lstlisting}[captionpos=b,escapeinside={(*}{*)}, language=RuleListsLanguage,backgroundcolor = \color{RuleListsLanguageBackgroundColor}, basicstyle=\scriptsize, caption={Example rule list learnt on the \namedataset{Compas} dataset~\cite{angwin2016machine}. The binary prediction is whether the offendent will recidivate within two years or not.}, label=fig:rule-list]
if (*\texttt{Prior-Crimes}*)(*$\neq 0$*) then True
else if (*\texttt{Juvenile-Felonies}*) (*$\leq$3*) and (*\texttt{Juvenile-Crimes}*) (*$\notin  \llbracket   1,3  \rrbracket   $*) 
(*\qquad*) then False
else True
\end{lstlisting} 

Rule lists are provably more expressive than decision trees of comparable size~\cite{Rivest}, mainly because there can be any arbitrary overlap between the supports of the different rules within the rule list (since they are ordered), while the supports of the leaves of a decision tree are inherently disjoint. Hence, any decision tree with depth $d$ can be translated into an equivalent rule list using rules involving at most $d$ boolean conditions, while the opposite is not true. A direct consequence from an information theory perspective is that rule lists encode more information regarding their training data than decision trees of comparable size~\cite{ferry:hal-04189566}. Nevertheless, despite their advantage in terms of compactness, rule lists remain less studied than decision trees in the privacy literature, which is why we focus on them in this work. There are a few notable exceptions, although they consider different setups and have different objectives~\cite{thaler2012faster,daniely2019locally}. On the one hand, Daniely and Feldman~\cite{daniely2019locally} discuss the sample complexity of learning decision lists in a non-interactive local differential privacy context. On the other hand, Thaler et al.~\cite{thaler2012faster} consider the problem of privately releasing a database whose rows consist of pre-defined decision lists, in the context of privacy-preserving data publishing.

Rule lists can be built either with exact methods (producing certifiably optimal models) such as the \texttt{CORELS}~\cite{corels_2} tree-based algorithm or with heuristic approaches~\cite{imodels2021}, which we specifically consider in this paper.
More precisely, we chose to focus on greedy learning algorithms (as introduced in Section~\ref{subsec:art-GRL-gini}), because this framework encompasses a wide range of interpretable models beyond rule lists, such as decision trees or random forests. This makes our results easily applicable to these other models. 

\subsection{Differential Privacy}
\label{subsec:background:dp}
Differential privacy (DP) is a mathematical property that yields strong privacy guarantees with respect to queries or computations performed on a database~\cite{foundations_DP}. The underlying idea is the following: an algorithm is differentially private if its distribution over outputs does not change much after adding or removing one sample (corresponding to one individual with personal information). % \textit{e.g.}) 
The objective of privacy-preserving machine learning is to reconcile two antagonist purposes: extracting useful correlations from data without revealing private information regarding any single individual. 
In particular in machine learning, DP can be integrated into the learning algorithm to ensure that the resulting model does not leak too much information with respect to the input dataset. 
In this context, a differentially private learning algorithm ensures that the distribution over outputs (\emph{i.e.}, possible models) is not impacted significantly by the addition or removal of a sample from the training set.

More formally, two datasets $\dataset$ and $\dataset'$ are said to be neighbouring if they differ at most by one sample, denoted by $\norm{1}{\dataset-\dataset'}\leq 1$ for $(\dataset,\dataset') \in \databases$ in which $\universe$ is the finite set of all possible samples in a dataset. Similarly, the number of elements in a dataset $\dataset$ is $\norm{1}{\dataset}$.
More details regarding these definitions and notations are provided within the Appendix~\ref{appendix:dist-db}. 

An algorithm $\algo : \mathbb{N}^{|\universe|} \mapsto \mathscr{Y}$ is $(\varepsilon, \delta)$-differentially private if $\forall~S~\subseteq~\mathscr{Y}, \forall~(\dataset,\dataset')~\in~(\mathbb{N}^{|\universe|})^2, ||\dataset-\dataset'||_1 \leq 1 $, we have:
$$ \p(\algo(\dataset) \in S)  \leq \exp (\varepsilon) \p(\algo(\dataset') \in S) + \delta\text{\cite{foundations_DP}.}$$

The parameter $\varepsilon$ controls the level of privacy of the algorithm as it defines how much the probability of an output can vary when adding or removing a sample. 
Typically, $\varepsilon=1$ is considered a reasonable value in terms of provided protection~\cite{choice_epsilon}. \emph{Pure DP} refers to when $\delta = 0$, while \emph{approximate DP} corresponds to values of $\delta>0$. Indeed, $\delta$ can be interpreted as a probability of ``total privacy failure". One possible instance of this $\delta$-failure could be that with probability $1-\delta$, the model will behave like pure DP while with probability $\delta$ (\emph{i.e.}, the failure probability), there will be no privacy guarantees at all. $\delta \ll \frac{1}{\norm{1}{\dataset}}$ is considered to be an absolute requirement since a $\delta$ of order $\mathcal{O}(1/\norm{1}{\dataset})$ could enable the total release of some samples of the dataset.

%tim : mis en commentaire pour gagner de la place
%
%ulr:remove newline
Intuitively, differentially private mechanisms often revolve around the idea of adding noise of a magnitude of order close to how steep the output function can change with slight variations of the input. 
More precisely, for a given function $f \in \mathbb{R}^k$,  its \textit{global sensitivity} precisely quantifies how much the value of $f$ can differ between any two neighbouring datasets.
%
%The global sensitivity of $f$ is denoted by $\Delta_p f$, in which $l_p$ stands for the
%%and can be computed as stated in the two following definitions {with 
% $l_1$ or $l_2$ norms. 
%\mjo{A discuter : function versus algorithm ?}
%\tim{Un algo est une succession de fonctions à mon sens.}
Formally, the global sensitivity of a function $f: \mathbb{N}^{|\universe|} \mapsto \mathbb{R}^k$ for norm $l_p$ (usually $l_1$ or $l_2$), denoted by $\Delta_p f$, is defined
for any neighbouring datasets $\dataset$ and $\dataset'$ as follows:
% \begin{definition}
%    $l1$-sensitivity of a function $f: \mathbb{N}^{|\universe|} \mapsto \mathbb{R}^k$ as:
$$\Delta_p f = \max_{\substack{\dataset,\dataset' \in \mathbb{N}^{|\universe|} \\ ||\dataset-\dataset'||_1 = 1 }} ||f(\dataset)-f(\dataset') ||_p.$$
% \end{definition}
%And  
%\begin{definition}
%$l2$-sensitivity of a function $f : \mathbb{N}^{|\universe|} \mapsto \mathbb{R}^k$ as:
%$$\Delta_2 f = \max_{\substack{\dataset,\dataset' \in \mathbb{N}^{|\universe|} \\ ||\dataset-\dataset'||_1 = 1 }} ||f(\dataset)-f(\dataset') ||_2$$
%\end{definition}
%ulr:remove newline

One of the shortcomings of global sensitivity is that it does not take into account the position in the latent space of the considered datasets. Nevertheless, it can effectively be used to obtain thorough DP guarantees.
%\mjo{introduire le terme global sensitivity ?}
A  straightforward approach to achieve DP is by adding noise to the output of the given function $f$.
%a given function $f \in \mathbb{R}^k$. \newline
Two popular differentially private mechanisms~\cite{foundations_DP} are based on this principle: the \textit{Laplace mechanism} $\mechan_{LAPLACE}^{\Delta_1}(\dataset,f,\varepsilon)$ and the \textit{Gaussian mechanism} $\mechan_{GAUSS}^{\Delta_2}(\dataset,f,\varepsilon, c)$. 
%\mjo{autres réfe ? ou juste celle là ? je les ai retiré ensuite}
%\cite{foundations_DP} \textit{e.g.}\jul{Add ref for the proofs (eg le livre ou programming dp ou les articles contenant les preuves si ils les citent)}
In these mechanisms, for each component $j$ of the function $f$, the amplitude of the added noise $N_j$ is scaled to its global sensitivity. %^{\Delta_p}
%$\Delta_p f$: 
%$\mechan_{NOISE}^{\Delta_p}(\dataset,f,\varepsilon) \coloneqq f(\dataset) + (\noise^{\Delta_p}_1, \dots, \noise^{\Delta_p}_k)$.\\ 
 %The most common differentially private mechanisms are the \textbf{Laplace Mechanism}, the \textbf{Gaussian Mechanism} and the \textbf{Exponential Mechanism}. 
%We will now review \cmjo{the main differentially private mechanisms.} 

 %\mjo{This category: au singulier ? Exponential mechanims and noise ?}
 %\jul{Exponential mechanism: plutôt ajout indirect de bruit, je rejoins mjo, la phrase est measleading...}
  %\mjo{its sensitivity using ....}\jul{Sensitivity precisely quantifies this notion, and can be computed as stated in the following definitions:}
%More precisely, t
The Laplace mechanism $\mechan_{LAPLACE}^{\Delta_1}(\dataset,f,\varepsilon)$ is based on $\Delta_1$ (global sensitivity with $l_1$ norm) along with Laplace noise: $\forall j \in \{1,\dots,k\}, \noise_j \sim Lap(\Delta_1 f/ \varepsilon)$. 
The probability density function of the Laplace distribution is $Lap(x \ | \ b) = \dfrac{1}{2b} \exp \big(\frac{-|x|}{b} \big)$ and the Laplace mechanism satisfies $(\varepsilon, 0)$-DP.

%\mjo{A voir notation pour densité de probabilité}

%For a given function $f$, the \textit{Laplace Mechanism} %~\cite{foundations_DP}}
%is defined as follows:
%$$\mechan_{LAPLACE}^{\Delta_1}(\dataset,f,\varepsilon) \coloneqq f(\dataset) + (\noise_1, \dots, \noise_k)$$
%where $\forall i \in \{1,\dots,k\}, \noise_i \sim Lap(\Delta_1 f/ \varepsilon)$ is the Laplace noise added to $f$. 
%The probability density function of the Laplace distribution is : $Lap(\dataset|b) = \dfrac{1}{2b} \exp \big(\frac{-|\dataset|}{b} \big)$. 
%The \textit{Laplace Mechanism} as defined above is $(\varepsilon, 0)$-DP.  %\newline 

The Gaussian mechanism $\mechan_{GAUSS}^{\Delta_2}(\dataset,f,\varepsilon, \delta)$ satisfies $(\varepsilon, \delta)$-DP. 
It uses $\Delta_2$ (global sensitivity with $l_2$ norm) together with Gaussian noise:\\ $\forall j \in \{1,\dots,k\}, \noise_j \sim  \mathscr{N} \Big (\mu = 0, \sigma = \frac{c \cdot \Delta_2 f}{\varepsilon} \Big)$ with $c^2 > 2 \log(\frac{1.25}{\delta})$.

%Using the $l2$-sensitivity for $f$, it is possible to show that the \textit{Gaussian Mechanism} is $(\varepsilon, \delta)$-DP:
%$$\mechan_{GAUSS}^{\Delta_2}(\dataset,f,\varepsilon, c) \coloneqq f(\dataset) + (\noise_1, \dots, \noise_k)$$
%where $\forall i \in \{1,\dots,k\}, \noise_i \sim  \mathscr{N} \Big (\mu = 0, \sigma = \dfrac{c \cdot \Delta_2 f}{\varepsilon} \Big)$ is the Gaussian noise added to $f$, with $c^2 > 2 \log(\frac{1.25}{\delta})$. % %\\

%\mjo{pour gagner de la place, j'imaginais qu'on pouvait avoir une présentation du type : (1) sensitivity (2) bruit (Cauchy, Gauss, Laplace) (3) DP Mechanisms associant certains bruits et certaines sensitivités avec les propriétés de DP associées}
%\jul{OK avec la proposition de MJo - aussi revoir la transition de noise addition mecanisms à sampling mecanisms}

%We now review 
%In contrast, 
Another common DP mechanism, called the \textit{Exponential mechanism} $\mechan_{EXP}^{\Delta u}(\dataset,u,\utilrange)$,  considers a (discrete) set of possible outputs $\utilrange$ and randomly samples one of them with respect to a utility metric. Thus in this mechanism, the noise added to comply with DP is introduced through a random sampling of the candidate outputs rather than added on the outputs themselves.
More precisely, let $ u : (\dataset,v) \mapsto u(\dataset,v)$ denote the utility function of element $v$ with respect to dataset $\dataset$. 
%We introduce 
The global sensitivity of this utility function is defined as: 
$$\Delta u = \max_{v\in \utilrange} \max_{\substack{\dataset,\dataset' \in \mathbb{N}^{|\universe|} \\ ||\dataset-\dataset'||_1 \leq 1}} |u(\dataset,v) - u(\dataset',v)|$$

The Exponential mechanism
$\mechan_{EXP}^{\Delta u}(\dataset,u,\utilrange)$ samples an element $v \in \utilrange$ with probability $p \propto \exp(\frac{\varepsilon \cdot u (\dataset,v)}{2 \cdot \Delta u})$ and satisfies $(\varepsilon,0)$-DP~\cite{foundations_DP}. 

%denoted $\mechan_{EXP}^{\Delta u}(\dataset,u,\utilrange)$ 
Finally, we will also use the \textit{Noisy Max Report} mechanism~\cite{foundations_DP}, which satisfies $(\varepsilon, \delta)$-DP. Considering a noisy mechanism  $\mechan_{noisy}$  satisfying $(\varepsilon, \delta)$-DP, the Noisy Max Report mechanism returns: 
$ \argmax_{v \in \mathcal{V}} \mechan_{noisy}(\dataset,u(\cdot, v),\varepsilon).$
%The \textit{Noisy Max Report} satisfies $(\varepsilon, \delta)$-DP~\cite{foundations_DP}. 
%, with possibly $\delta=0$ depending on the privacy guarantees of the underlying noisy mechanism used.

Among others, DP comes with two fundamental properties. First, the \textit{post-processing} property ensures that DP guarantees are not affected by post-processing the output of a DP mechanism. Second, the \textit{composability} property enables the composition of different differentially private mechanisms (sequentially or in parallel) and the computation of the global privacy budget. In a nutshell, sequential composition states that if several DP mechanisms are applied on overlapping datasets, their privacy budgets sum up, whereas parallel composition considers the application of several DP mechanisms on disjoints datasets, in which case there is no need to sum up their budgets. More details regarding these properties are provided in the Appendix~\ref{appendix:comp_post_dp}.
\section{Building Blocks}
\label{sec:art}
%Seb: j'ai mis en commentaires l'introduction pour gagner de la place
%\seb{je pense qu'on devrait changer le titre de la section peut être à building blocks plutôt que related work}
%In this section, we introduce the different building blocks that are necessary for the design of our framework. 
%More precisely, 
%We first describe the greedy algorithm %that we consider to consider 
%as the baseline for learning rule lists models as well as the computation of the Gini impurity index. 
%We then review the notion of smooth sensitivity, before showing how it can be used to get DP guarantees.
%\seb{à voir si on souhaite présenter la Smooth sensitivity directement après la partie DP}

\subsection{A Greedy Algorithm for Learning Rule Lists}
\label{subsec:art-GRL-gini}
%Greedy algorithms are defined as algorithms that select the best available option at \cmjo{a given step} time $t$ without taking into account whether or not it would lead to a sub-optimal result at \cmjo{a the next step.} time $t+1$. 
%\mjo{commencer par donner l'idée générale d'un glouton pour les rule lists}

Greedy algorithms are widely used for learning decision tree models. 
For instance, the commonly used \texttt{CART} algorithm~\cite{cart} iteratively builds a decision tree in a top-down manner, by successively selecting the feature (and split value) yielding the best information gain value according to some pre-defined criterion. 
While algorithms for learning rule lists in a greedy manner are far less popular than their counterparts for learning decision trees, some implementations exist in the literature. 
For instance, the \texttt{imodels} \footnote{\url{https://github.com/csinva/imodels}} 
library~\cite{imodels2021} contains algorithms for learning different types of interpretable models, including rule lists (denoted \GRL{}). 
More precisely, \GRL{} iteratively calls \cart{} to build a depth-one decision tree at each level of the rule list, optimizing a given information gain criterion. 
%Seb: j'ai mis en commentaire pour gagner de la place
%Indeed, a rule list $\rlist = (\aruleind{1}, \dots, \aruleind{K}, \aruleind{0})$ with $\aruleind{i} = p_i \rightarrow q_i$ can be represented by a binary tree with two types of nodes: decision nodes associated to rule's antecedent $p_i$ and leaf nodes associated to its prediction $q_i$. 
%Thus, each decision node $p_i$ leads to two children: a leaf node $q_i$ if the antecedent $p_i$ is true or the next decision node $p_{i+1}$ otherwise.
Just like for decision trees, greedy algorithms for building rule lists successively select the best rule $\aruleind{i} = p_i \rightarrow q_i$ given some information gain criterion. 
%More precisely, 
Thus, at each level of the rule list being built, the \GRL{} algorithm iterates through all possible rules and keeps the one leading to the best information gain value.
%\seb{il faudrait préciser d'où vient l'algorithme GRL mentionné ci-dessus (c'est la première fois qu'on le mentionne)}

\subsection{Gini Impurity for Rule Lists}

%Throughout 
In this paper, we consider the Gini impurity index originally used in the \texttt{CART}~\cite{cart} algorithm as a measure of the information gain.
In a nutshell, this index quantifies how well a %node 
rule separates the data into two categories with respect to different labels, with the value of zero being reached when the examples are perfectly separated. 
The algorithm stops when all the samples are classified, but other stopping criteria can be implemented such as a maximum length on the list of rules or a minimum support condition on each rule (\emph{i.e.}, number of points left to be classified). 
%at this node). 
%What we call support of a node is the number of points left to be classified at this node, independent from how the split is eventually made. 
%In case of equality, prefer the term with the lowest $\mathscr{G}_c(r)$. 
%\mjo{Parler des conditions d'arrêt de \DPGRL{}: tous classifiés, max length, min support, valeur gini}

%\mjo{La suite de cette section est une contribution ? Si oui, rappeler d'abord comment on calcule de gini dans des arbres et pointer la différence avec des rule lists.}
%\tim{Non ce n'est pas une contribution c'est comme ça qu'on calcule dans les arbres et on a repris pareil pour les rl, cf réponse de julien.}
%\jul{Quel est le but de la suite de cette sous-section ? Introduire la formule mathématique du Gini pour les RLs ? Si oui je ne comprends pas trop pourquoi on parle d'abord d'arbres. Pour la question sur calcul, sur les arbres c'est la somme de droite et gauche pondérés par leurs supports relatifs je pense, et idem pour les RLs (y compris dans imodels, comme expliqué dans mon texte ils appellent CART (de sklearn) pour construire un arbre à 1 noeud à chaque étage de la RL). Calculer le gini uniquement sur la partie capturée serait faisable, mais j'avais essayé et ça laisse des supports trop "impurs" ce qui pénalise tout ce qui vient après et la perf globale de la RL.}
Consider a given rule $\aruleind{}$ for a specific node 
of a pre-existing list of rules, which means that some samples were already captured by previous rules and are not accounted for.
Let $\setcapt{r} \subset \dataset$ be the subset of samples captured by rule $\aruleind{}$, in which $\ncapt{r}$ is the number of samples in $\setcapt{r}$ and $\nleft{r}$ the number of samples not captured by rule $\aruleind{}$.
%$C(r) \subseteq \mathscr{S}$ the set of all samples. 
For a rule list $\rlist={(\aruleind{1}, \dots, \aruleind{K}, \aruleind{0})}$, and a given position $j$ in the sequence, let $\nremain{j}$ be the number of samples not captured by previous rules $\aruleind{1} \ldots \aruleind{j-1}$.
In particular, this means that $\nremain{j} = \ncapt{r_j} + \nleft{r_j} = \nsamples - \sum_{i=1}^{j-1}{\ncapt{r_i}}$.
%$\tilde{n}\coloneqq | \{s \in \mathscr{S} : \text{$s$ not captured by previous rules} \} | $
%\tim{J'ai remplacé tous les $j$ par des $r$ oudes $r_j$ pour que la notation suive}
%Let $n_r$ the number of samples captured by rule $r$, $n_r \coloneqq |C(r)|$.  
%Let $n_l$ the number of samples not captured by rule $r$. 
%Note : $n_r + n_l = \tilde{n}  \neq |\mathscr{S}|$. 
In addition, let $\labelpred{r}{c}$ denote the average outcome (\textit{i.e.} the predicted label) of the rule $\arule$, 
%A sample $s \in \mathscr{S}$ can be represented as a tuple $s=(x_s,y_s)$ where $x_s$ encodes the features and $y_s$ the label. 
$\labelpred{r}{c} = \frac{1}{\ncapt{r}} \sum_{\asampleind{} \in C(r) } \alabelind{s}$. %\cmjo{(where a sample $\asampleind{i}$ is defined by $(\afeatind{i}, \alabelind{i})$}
Similarly, the average outcome of the remaining samples is $$\labelpred{r}{l} =   \frac{1}{\nleft{r}} \sum_{ s \in \dataset \setminus \big( \cup_{i=1}^{j-1} C(r_i) \ \cup \ C(r) \big)}\alabelind{s}$$

%\mjo{A voir pour simplifier la dernière égalité - }

The Gini impurity reduction with respect to rule $\arule$ is denoted as $\gini{\arule}$. 
It can be divided into two terms $\ginicapt{\arule}$ and $\ginileft{\arule}$, respectively for the samples caught and the ones not caught by the rule: 
$\gini{\arule} =\ginicapt{\arule} + \ginileft{\arule}$.
Note that we not only consider the samples caught by the rule (through $\ginicapt{\arule}$) but also those which are not (through $\ginileft{\arule}$) as it matters for the following rules in the rule list.
%One could argue that we should only account for the Gini impurity of the samples caught but adding the term $ \ginileft{\arule}$ is an improvement to the greedy heuristic to `ease' the following rules.
%\mjo{repartir de la forume du gini index pour les arbres ? $gini = 1 -sum_{i \in class} P(i)^2$ avec $P(i)$ la proba d'être dans la classe $i$ (nb avec label $i$ / nb total exemples du noeud)}
For binary classification, the Gini impurity reduction for a rule $\arule$ at %node 
position $j$ is given by:
%$\ginicapt{\arule} = \dfrac{\ncapt{\arule}}{\nremain{j}} \big( 1 - \sum_{i=1}^{\text{number of classes}} \p [ y_s = i]^2 \big)$.
%
%This formula simplifies for binary classification as :
\begin{align}
\label{eqn:gini_c}
  \ginicapt{\arule} &=  \dfrac{\ncapt{\arule}}{\nremain{j}} \big(1- \labelpred{\arule}{c}^2 - (1-\labelpred{\arule}{c})^2 \big)\end{align}

 \begin{align}
 \label{eqn:gini_l}
   \ginileft{\arule} &= \dfrac{\nleft{\arule}}{\nremain{j}} \big(1- \labelpred{\arule}{l}^2 - (1-\labelpred{\arule}{l})^2 \big)  
 \end{align}

%\mjo{A voir les formules : des tilde et des hat pour les labels ?}

\subsection{Smooth Sensitivity}
\label{subsec:art-smooth-sensitivity}
%\mjo{j'ai mis en commentaire les définitions, théorèmes, lemmes de cette partie (en supposant que ce n'est pas utilisé par la suite, j'espère que c'est ok ...)}
The DP mechanisms described in Section~\ref{subsec:background:dp} rely on the notion of global sensitivity.
However, some functions only display a very loose bound for their global sensitivity. 
For instance, the global sensitivity of the Gini impurity is $0.5$, irrespective of the actual number of samples left to be classified. This is considerably high, given that the Gini impurity takes values in $[0,1]$. 
%This issue arises with other functions too. 
To address this limit, previous work~\cite{nissim} 
proposed a way to compel a tighter bound on the added noise. Firstly, they have introduced the notion of %local sensitivity of a function $f$: 
%in the stead of its global sensitivity :
% \begin{definition}
% \textbf{Local Sensitivity}
%let $f : \mathbb{N}^{|\universe|} \mapsto \mathbb{R}^k$ and \cmjo{a dataset} $\dataset \in  \mathbb{N}^{|\universe|}$. 
the \textit{local sensitivity} of a function $f : \mathbb{N}^{|\universe|} \mapsto \mathbb{R}^k$ at a dataset $\dataset$, denoted $\ls_f(\dataset)$, as:
%
%$$\ls_f(\dataset)=  
$\max_{\substack{\dataset' \in \mathbb{N}^{|\universe|} : \\ ||\dataset-\dataset'||_1 = 1 }} ||f(\dataset)-f(\dataset')||_1$.     
% \end{definition}
However, replacing directly the global sensitivity by local sensitivity does not yield privacy guarantees. 

This motivated the formulation of a refined sensitivity notion denoted as \emph{smooth sensitivity}.
This notion exploits a \emph{smooth upper bound} of $\ls_f(\dataset)$, denoted by $\smoothub_{f,\beta}(\dataset)$, as follows. 
For $\beta >0$, $\smoothub_{f,\beta}(\dataset) : \mathbb{N}^{|\universe|} \mapsto \mathbb{R}^+$ is a $\beta$-smooth upper bound on the local sensitivity of $f$ if it satisfies :    
\begin{align}
    &\forall \dataset \in \mathbb{N}^{|\universe|}, \forall \dataset' \in \mathbb{N}^{|\universe|} \ s.t. \ ||\dataset-\dataset'||_1 = 1, \nonumber \\
     & \smoothub_{f,\beta}(\dataset) \geq \ls_f(\dataset) \quad \text{and}  \quad  \smoothub_{f,\beta}(\dataset) \leq e^{\beta} \smoothub_{f,\beta}(\dataset') \label{eq_smooth}
\end{align}
%\end{definition}
%\mjo{notation : faire apparaitre $f$ dans $\smoothub_{\beta}$ ? equation dans l'autre sens pour $e^{-\beta}$ ? pour ne pas noter  la smooth senstisivy par $\Gamma_{1,\beta}^* f(\dataset)$
%\mjo{A voir entre smooth UB of LS et smooth sensitivity}
%\vspace{0.5cm}
%\begin{theorem}
The smallest function to satisfy Equation~(\ref{eq_smooth}) is called the \textit{smooth sensitivity} and denoted $\smooth_{f,\beta}(\dataset)$:
\[\text{For } \beta >0, \smooth_{f,\beta}(\dataset) = \max_{\dataset' \in \mathbb{N}^{|\universe|}} \ls_f(\dataset') e^{-\beta ||\dataset-\dataset'||_1} \]
%\end{theorem}

%\mjo{Ecrire des questions n'est pas très habituel}

Nissim et al.~\cite{nissim} proposed an iterative computation of the smooth sensitivity (Lemma~\ref{lemma-smooth}) considering datasets
%Define $\mathcal{T}_k$ the local sensitivity of $f$ at distance $k$ (we consider databases 
than can vary up to $k$ samples rather than $1$. 
Let $\mathcal{T}_k$ denote the local sensitivity of $f$ at distance $k$: 
$$\mathcal{T}_k(\dataset) =\max \big\{LS_f(\dataset') \ \big| \ ||\dataset' - \dataset||_1 \leq k \big\}.$$ 
%The proof of this lemma is recalled in Appendix~\ref{appendix:lemma-smooth-iterative-computation}.

%One can then evaluate $\smooth_{f,\beta}(\dataset)$ using $\mathcal{T}_k(\dataset)$ as follows.
\begin{lemma}%[Iterative Computation of $\smooth_{f,\beta}$]
\hspace{-2pt}\normalfont{\textbf{\cite{nissim}}}
$ \smooth_{f,\beta}(\dataset) = \max \big\{e^{-\beta k} \mathcal{T}_k(\dataset) \ \big | \ k \in \mathbb{N}\}$. (proof recalled in Appendix~\ref{appendix:lemma-smooth-iterative-computation}) 
\label{lemma-smooth}
\end{lemma}

%\seb{il faudrait détailler un peu plus le lemme ci-dessus}

As stated by~\cite{dpsmooth-forest, Zafarani2020,Sun2020}, smooth sensitivity is a very powerful tool to replace global sensitivity
%DP mechanisms based on smooth sensitivity generally incurs an higher accuracy than global sensitivity 
for differentially private machine learning.
%Although smooth sensitivity is a very powerful tool to replace global sensitivity, 
%%and \ctim{thus improve on the utility}, 
%it comes with its own challenges. 
%First, 
However, finding a closed form for $\smooth_{f,\beta}(\dataset)$ is difficult and sometimes requires to make stronger assumptions on the model. 
%In practice, smooth sensitivity is rarely used due to the challenge in implementing it.
%\mjo{une petite phrase de transition ?}
%\subsection{Differentially Private Mechanisms with Smooth Sensitivity}
%\mjo{Ajouter une petite phrase avant de commencer. Ajouter référence pour ces méchanismes avec du bruit de Cauchy ou du bruit de Laplace ?}
Nonetheless, two DP mechanisms were proposed by~\cite{nissim} based on the smooth sensitivity.

The first one is based on Cauchy noise %$\mechan_{CAUCHY}^{\smooth_{\cdot, \beta}}$ 
and uses an additional parameter $\gamma$:
$$\mechan_{CAUCHY}^{\smooth_{\cdot, \beta}}(\dataset, f, \varepsilon):\dataset \mapsto f(\dataset) + \dfrac{2(\gamma+1)\smooth_{f,\beta}(\dataset)}{\varepsilon} \cdot \eta$$
with $\beta \leq \frac{\varepsilon}{2(\gamma +1)}$, $\gamma > 1$ and $\eta \sim h(z) \propto \frac{1}{1+|z|^{\gamma}}$ the Cauchy noise. This mechanism satisfies $(\varepsilon$,0)-DP.\newline

The second one uses Laplace noise and satisfies $(\varepsilon, \delta)$-DP:
$$\mechan_{LAPLACE}^{\smooth_{\cdot,\beta}}(\dataset, f, \varepsilon, \delta): \dataset \mapsto f(\dataset) \ + \  \dfrac{2 \cdot \smooth_{f,\beta}(\dataset)}{\varepsilon} \cdot \eta  $$  with $\beta \leq \frac{\varepsilon}{2 \log(2/\delta)}$ and 
$\eta \sim Lap(1)$, the Laplace noise. 
%\rqe{ajouter $\delta$ dans la définition de $\mechan_{LAPLACE}$} \ctim{DONE}. 
%
%We denote them $\mechan_{CAUCHY}^{\smooth_{\cdot, \beta}}$ and $\mechan_{LAPLACE}^{\smooth_{\cdot,\beta}}$ and they are respectively $(\varepsilon$,0)-DP and $(\varepsilon, \delta)$-DP.  %and $\smooth_{f,\beta}$}. 
%
%\seb{il faudrait expliquer chacun de ces mécanismes et expliquer leur intuition en une ou deux phrases}
%$\smblksquare$ Let $\beta \leq \frac{\varepsilon}{2(\gamma +1)}\ , \ \gamma > 1\ , \  \eta \sim h(z) \propto \frac{1}{1+|z|^{\gamma}}$ $\mechan_{CAUCHY}^{\smooth_{\cdot, \beta}}(\dataset, f, \varepsilon):\dataset \mapsto f(\dataset) + \dfrac{2(\gamma+1)\smooth_{f,\beta}(\dataset)}{\varepsilon} \cdot \eta$ 
%
%$\smblksquare$ Let $\beta \leq \frac{\varepsilon}{2 \log(2/\delta)} \ , \ \eta \sim Lap(1)$, $\mechan_{LAPLACE}^{\smooth_{\cdot,\beta}}(\dataset, f, \varepsilon): \dataset \mapsto f(\dataset) + \dfrac{2 \cdot \smooth_{f,\beta}(\dataset)}{\varepsilon}$
%
%\mjo{A définir ces 2 mechanims comme dans le background}
%\mjo{il y a un $\delta$ et un $\gamma$ dans le 2e mec. ?}

Note that in contrast to global sensitivity, adding Laplace noise within the framework of smooth sensitivity %adding Laplace noise 
does not yield pure DP anymore but approximate one. 
Furthermore, to obtain pure DP guarantees along with smooth sensitivity, heavy-tailed noise distributions must be considered, such as the Cauchy distribution.
In the following, we fix $\beta = \frac{\varepsilon_{node}}{2\log(2/\delta_{node})}$ for the $\beta$-smooth upper bound.
\section{A differentially private Greedy Learning Algorithm for Rule Lists}
\label{sec:contrib}
%In this section, we %\mjo{A discuter "more complex sensitivity denoted as smooth sensitivity" 
%- notation avec une lettre grecque ? par exemple pour $LS_f(\dataset)$ utiliser $\delta_1 f(\dataset)$. 
%- Est-ce qu'on a bien $\Delta_1 f = \max_{\dataset' \in \mathbb{N}^{|\universe|}} \delta_1 f(\dataset')$ ?}
%\begin{definition}
We now introduce our framework for learning differentially private rule lists leveraging smooth sensitivity. 
Unlike \cite{dpsmooth-forest} who integrate smooth sensitivity to determine the majority class for a leaf in a tree, we integrate it to determine the rule with the best Gini impurity.
%More precisely, we 
%Seb: j'ai mis en commentaires pour gagner de la place
%We first demonstrate how to precisely compute the smooth sensitivity of the Gini impurity before leveraging it to design a differentially private \GRL{} algorithm. 

\subsection{Establishing the Smooth Sensitivity of the Gini Impurity}
\label{subsec:contrib-smooth-gini}

%\cite{local-sensitivity} have characterized 
The local sensitivity for the Gini impurity has been characterized in~\cite{local-sensitivity}. Considering the support $\nremain{j}$ of the $j$th rule, it is defined by: 
%Recall that at a given node $j$ of the rule list, the support is given by $\nremain{j} = \nleft{r_{j-1}}$ in which $r_{j-1}$ is the rule added at step $j-1$ to the rule list.
%\seb{à valider si on la mentionner auparavant et voir si on veut vraiment le répéter}
%\mjo{simplifier car dit avant je pense}
$$ LS_\mathcal{G}(\nremain{j}) = 1 - \Big( \dfrac{\nremain{j}}{\nremain{j}+1} \Big)^2 - \Big( \dfrac{1}{\nremain{j}+1} \Big)^2$$

%\mjo{reprendre les notations de la section \ref{subsec:art-GRL-gini} pour $D$. A un noeud donné $\arule$: on avait noté par $\nremain{r}$ le nombre d'exemples restants (dire que c'est le support ?).}
%Pour local sensitivity c'est actuellement $LS$ (et pourquoi pas $\delta$

Given a minimal support $\minsupp$ imposed %on each node, 
for each selection of rule (\emph{i.e.,} a minimum number of samples that a rule must capture in order to be considered for inclusion within the built rule list), we have derived in Theorem~\ref{theorem-smooth} %an exact %heuristic 
a method to compute the smooth sensitivity of the Gini impurity.
%\cmjo{method} to compute the smooth sensitivity of a node in a rule list, given the minimal support $\minsupp$ imposed 
%on the node.  
%\cmjo{on each node for the DP$-\GRL$ algo}

\begin{theorem}[Smooth Sensitivity of the Gini impurity]
\label{theorem-smooth}
Let $\minsupp \in \mathbb{N}^*$ be the given minimum support. %as defined above.  
By inverting the parameter $k$ and the variable $\dataset$ in the function $\mathcal{T}_k(\dataset)$, we define the following function : 
\[\xi_{\dataset,\beta}(k): \left|   \begin{array}{ccl}
    \mathbb{N} & \longrightarrow  & \mathbb{R}^+\\
    k & \longmapsto & e^{-k \beta} \cdot g \big[ \max(\minsupp, \norm{1}{\dataset}-k) \big]  \\
  \end{array} \right. \]
in which  \[g: \left|   \begin{array}{ccl}
    \mathbb{R}^+ & \longrightarrow  & [0,1] \\
    x & \longmapsto & 1 - \Big( \dfrac{x}{x+1} \Big)^2 - \Big( \dfrac{1}{x+1} \Big)^2 \\
  \end{array} \right. \]

The smooth sensitivity of a rule with a dataset $\dataset$ of points that remain to classify is : 
%\mjo{why support $\dataset$?} 
%\tim{la smooth sensitivity prend en compte le support du noeud et le support mininimal!}
%\mjo{oui, c'est la notation $\dataset$ qui m'embête}
%\tim{a la base c'était x. Disons que le support peut tout autant désigner l'ensemble de points qu'il reste à classifier, que le cardinal de cete ensemble.}
%\mjo{OK, c'est un remplacement que j'ai fait trop rapidement alors ... C'est plus clair de préciser que c'est sur l'ensemble de points courant :-)}
%\tim{ok}
$$ \smooth_{ \mathcal{G}, \beta}(\dataset) =  \max\Big[\xi_{\dataset,\beta}(0),\ \xi_{\dataset,\beta}(\floor{t}),\ \xi_{\dataset,\beta}(\ceil{t}),\ \xi_{\dataset,\beta}(\norm{1}{\dataset}-\minsupp) \Big] $$
with $t = \norm{1}{\dataset} - \dfrac{1-\beta -  \sqrt{ (1-\beta)^2 - 4 \beta} } {2 \beta}$ if well defined and otherwise $0$.
\begin{proof}
The detailed proof is provided in Appendices~\ref{appendix:smooth-sensitivity-1} and~\ref{appendix:smooth-sensitivity-lambda} in which we first prove it for $\Lambda = 1$ and generalize the proof for $\Lambda \in \mathbb{N}^*$. 
Crucially, recall that the smooth sensitivity is the same for any rule at a given position since we have proven that the smooth sensitivity of the Gini impurity only takes into account the number of elements left to be classified (and not how the rule captures them or not).
    The proof is a proof by exhaustion, in which the smooth sensitivity is computed as $\smooth_{ \mathcal{G}, \beta}(x) = \max_{k \in \mathbb{N}} \ e^{-k \beta} \mathcal{T}_k(x)$. 
We first determine the function $\mathcal{T}_k(x)$ in which $x$ is a dataset. 
We observe that it does not depend on the actual value of the dataset but solely the number of samples it contains. 
We obtain $\mathcal{T}_k(x) = g\left(\text{max}(1,||x||_1-k)\right)$. 
Given that we managed to obtain a closed form for $\mathcal{T}_k(x)$ in which $k$ directly intervenes, we now consider $x$ to be a parameter and put $k$ as a variable of our function hence the introduction of function $\xi_{x,\beta}(t) =  \ e^{-k \beta} \mathcal{T}_k(x)$. 
Rather, we study this function on $\mathbb{R}^+$ since it is differentiable. 
Through the cancellation of the derivative, we are able to find the minima and the maxima of the function. 
However, since these are $\mathbb{R}$-valued maxima, we finally truncate them to the closest higher and inferior integers to obtain the smooth sensitivity. 
Note that since we associate the sign of the derivative to a polynomial, it gives us extra control over the monotony of the function $\xi$ since we know a polynomial takes the sign of the highest degree coefficient outside of the roots (granted that they exist) and the opposite inside the roots. 
For certain values of $\beta$ such maxima may not exist because they are computed as the roots of a polynomial whose values depend on $\beta$. 
For this reason, we state that the formula should encompass $t$ only if it is well defined (\emph{i.e.}, the value inside the square root is not negative) and otherwise replace it by 0. 
Proving that the smooth sensitivity only depends on $||x||$ is a core result of our approach. 
Thanks to this, at each iteration we only need to query the number of elements left to classify and apply the same smooth sensitivity to all rules (the Gini impurity depends on the split made by the rule but its smooth sensitivity is independent of it).
\end{proof}
\end{theorem}

\begin{figure}[h!]
    \centering
    \includegraphics[width=0.98\linewidth] {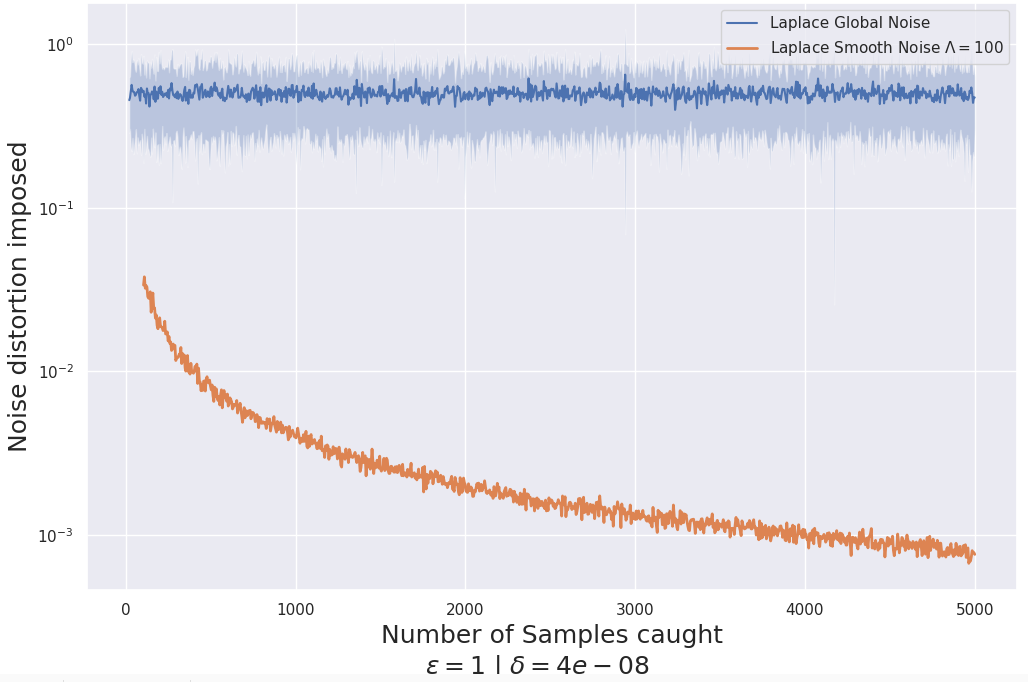}    %1.1
    \caption{Comparison of the amplitude (log scale) of the noise added by the Laplace mechanism scaled to either the Smooth or Global Sensitivities.}
    \label{fig:sensitivity}
\end{figure}

Figure~\ref{fig:sensitivity} gives an overview on the amount of noise one has to add to the computed Gini impurity to get a target DP guarantee, using either global or smooth sensitivity. % illustrates how smooth sensitivity leads to a more calibrated noise than global sensitivity. 
More precisely in this figure, we display the noise distortion generated for a fixed $\varepsilon=1$ by each DP mechanism as a function of the number of samples captured by the rule.
Importantly, we observe how the use of smooth sensitivity allows to scale down the generated noise when considering more samples. This is not the case for global sensitivity, which is dataset-independent. Overall, the noise added using smooth sensitivity is far inferior to the noise generated by a global sensitivity mechanism for a similar level of a privacy - which is a promising preliminary result for the implementation of a private model relying on the smooth sensitivity of the Gini Impurity. 

Many learning algorithms use a regularization parameter scaling with the size or complexity of the model to enhance interpretability and reduce overfitting~\cite{regularization}. 
In our case, apart from being a key factor for the smooth sensitivity computation, the minimum support also leads to a better comprehensibility of the resulting rule list by encouraging its sparsity, as there can only be as many as $\frac{1}{\lambda}$ rules where $\lambda$ is the relative minimum support defined as $\lambda = \frac{\Lambda}{\norm{1}{x}}$. It also plays the role of a regularization parameter as it helps the model to not overfit. Note that other types of rule lists learning algorithms, such as the \texttt{CORELS} exact method, also consider a regularization term based on the number of rules within the built rule list.

\subsection{Greedy Learning of Differentially Private Rule Lists Leveraging Smooth Sensitivity}
\label{subsec:dp-greedy-rl}

The proposed differentially private algorithm for learning rule lists, called \DPGRL, is detailed in Algorithm~\ref{alg:smooth-dp}.
This algorithm is based on the smooth sensitivity established in the previous subsection and uses Laplace noise. It takes as input a set of rules $\ruleset$ that is assumed to be publicly known and is not obtained as a computation from the data. Note that this assumption is consistent with the literature. For a set of parameters (Gini impurity computation, rule list size and minimum support, and privacy guarantees) this greedy algorithm iteratively adds a new rule to the rule list $RL$. 
At each step, it checks whether the support in the current remaining dataset $X_{rem}$ verifies the minimum support condition (Line~\ref{line:cond-min-sup}), including the confidence threshold computed once for all (Line~\ref{line:confidence}). %(see Algorithm~\ref{alg:threshold})
%The main loop (from line~\ref{line:begin-while} to~\ref{line:end-while}) 
For each rule $r \in \ruleset$, its noisy Gini impurity is computed using our proposed Laplace mechanism based on smooth sensitivity at Line~\ref{line:gini-smooth-laplace} and
%consists in computing the Gini Impurity of each rule in $\ruleset$ and returning
the rule $R^\bigstar$ with lowest noisy Gini is returned. 
%\mjo{le fait d'avoir un ensemble de règles précalculées $\ruleset$ pourrait poser problème pour la DP ?}
%\tim{non. On considère que ça fait partie des données, fixées de départ.}
%\seb{je confirme si c'est connu au départ (et non appris sur les données) ça ne pose pas de problème}
%\mjo{On en discute lundi ? }
%
$R^\bigstar$ is then added to $RL$ %the rule list 
with its DP prediction $q^\bigstar$ (Line~\ref{line:prediction}) and removed from $\ruleset$. 
The main loop is stopped when (1) the rule list reaches the maximum length, (2) the support condition is not verified anymore or (3) adding a rule does not improve the Gini impurity value. 

%is parametrized by the minimum support threshlod $\minsupp$, the maximum length of a rule list $K$, a privacy budget $\varepsilon$ and $\delta$, and a confidence value $\mathcal{C}$. 

%\mjo{Commencer par décrire l'algo~\ref{alg:smooth-dp} - ajouter numérotation des lignes pour aider ? } 
%\tim{fait}

%\mjo{conserver $c_0$ et $c_1$ de l'output ?  Trouvé les package algo fournis étaient trop vieux ... }

\begin{algorithm}[h!]
\caption{Approximate $(\varepsilon, \delta)-$DP-Greedy Rule List with Smooth Sensitivity}
\label{alg:smooth-dp}
%\begin{algorithmic}[1]
   %\STATE 
   {\bfseries Input:} Dataset $\dataset \in \databases$, Rule set $\ruleset$ \\
   %\STATE 
   {\bfseries Parameters:}  Min support of the dataset $\lambda$, Max length of a rule list $K$, DP budget $(\varepsilon, \delta)$,  Confidence $\mathcal{C}$ \\
   %\STATE
   {\bfseries Output:} Rule List $RL$ (and noisy counts $c_0, c_1$) 
   %\hspace*{\algorithmicindent} 
   
\begin{algorithmic}[1]
   \STATE $X_{rem} \gets\dataset $, $R_{rem} \gets \ruleset$, $RL \gets []$, \quad   \COMMENT{Initialisation} \\
   $\Lambda \gets \lfloor \norm{1}{x} \times \lambda \rfloor$, \quad  $\textup{Stop} \gets \textup{False}$  \\
   \STATE $\mathcal{T} \gets \textbf{\texttt{confidence\_threshold}}(\mathcal{C})$ 
 \label{line:confidence} \\

 \WHILE{$RL\cdot \textbf{\texttt{size}}() < K \hspace{0.5em} \AND \hspace{0.5em} \neg \textup{Stop}$} \label{line:begin-while}
    \IF{$ \mathcal{M}_{LAPLACE}^{\Delta_1}(X_{rem}, count(X_{rem}, \cdot),
    %\norm{1}{\cdot}, 
    \varepsilon_{node}) < \Lambda + \mathcal{T}$} \label{line:cond-min-sup}
    \STATE $\textup{Stop} \gets \textup{True}$ \label{algo-test}\\
    \ELSE    
        \STATE $\mathcal{G}_{bound} \gets \mathcal{M}_{LAPLACE}^{\smooth_{\cdot, \beta}}(\emptyset, \mathscr{G}_{X_{rem}}(\cdot), \varepsilon_{node}, \delta_{node})$ \label{line:gini-smooth-laplace-init}
        \STATE $\mathcal{G}^{\bigstar} \gets \mathcal{G}_{bound} $ \COMMENT{no rule added gini}  
        \STATE $R^{\bigstar} \gets \emptyset $, $q^{\bigstar} \gets \textbf{\texttt{pred\_DP}}(\emptyset, X_{rem})$ 
    \FOR{$r \in  R_{rem}$} \label{line:loop-rule}
        \STATE $ \mathcal{G} \gets \mathcal{M}_{LAPLACE}^{\smooth_{\cdot, \beta}}(r, \mathscr{G}_{X_{rem}}(\cdot), \varepsilon_{node}, \delta_{node})$ \label{line:gini-smooth-laplace}\\
        \IF{$\mathcal{G} < \mathcal{G}^{\bigstar}$}
            \STATE $\mathcal{G}^{\bigstar} \gets \mathcal{G}$, $R^{\bigstar} \gets r$\\ 
            \STATE $q^{\bigstar} \gets \textbf{\texttt{pred\_DP}}(r, X_{rem})$ \label{line:prediction}\\
        \ENDIF         
   \ENDFOR 
    \IF{$R^{\bigstar} = \emptyset$}
        \STATE $\textup{Stop} \gets \textup{True}$\\
    \ELSE
        \STATE $RL\cdot \textbf{\texttt{append}}(R^{\bigstar}, q^{\bigstar})$\\
        \STATE $\textbf{\texttt{update$_{DB}$}}(X_{rem} \gets X_{rem} \setminus \mathcal{C}(R^{\bigstar}))$\\
    \ENDIF
    
   \ENDIF
   \ENDWHILE \label{line:end-while}
\end{algorithmic}
\end{algorithm}

%\tim{pourquoi la notation infinie? dans l'algo?}
\textbf{DP computation of the rules' predictions (Line~\ref{line:prediction}).}
In Algorithm~\ref{alg:smooth-dp}, it is necessary to determine the prediction for each rule in a differentially private manner. 
%In \cmjo{line xxx of the} algorithm, 
Indeed, in the non-DP setup, the prediction is computed as the majority class among the samples caught by the rule. 
%(\emph{i.e.}, $q = \ind[\hat{y}_r \geq 0.5]$). 
%\mjo{il manque la macro pour l'indicatrice}
However, such a deterministic selection of the best prediction is not compatible with DP.
For instance, consider two neighbouring datasets $\dataset$ and $\dataset'$. 
Let $\arule$ be a rule picked from the rule list built on $\dataset$. % and assume that the best prediction is \textbf{True} for $\arule$. %\jul{reformuler la phrase juste avant}
If $\dataset'$ is $\dataset$ deprived from one element that would
%fix $\arule$ to \textbf{True} (basically a positive sample caught by $\arule$) then if the best prediction for $r$ becomes \textbf{False} on $\dataset$
flip the outcome of $\arule$, the probability of observing this outcome in the built rule list is also flipped from $1$ to $0$ breaking any DP guarantee.
%outcome of the output has been changed from a positive probability to a null one, violating the DP guarantee. 
Thus, the rules' predictions have to be determined using DP-protected counts. 
In our implementation (Algorithm~\ref{alg:predDP}), we use the Laplace mechanism based on the global sensitivity to compute the counts for each rule that are later used to determine the rule's prediction.
%
%\rqe{donc pure DP et propriété de composition pour l'ensemble ? }
%\ctim{c'est ça}
%
Thus, Algorithm~\ref{alg:predDP} provides pure DP guarantees. Note that since counting queries have sensitivity $1$ while their output takes values in $[0,\nsamples]$, this is a reasonably low value and the mechanisms based on global sensitivity usually yield good utility in this setting.

\begin{algorithm}[h!]
\caption{ \textbf{Function \texttt{pred\_DP}} : }
\label{alg:predDP}
 %\STATE 
 {\bfseries Input:} Rule $r$, Remaining samples $X_{rem}$ \\
 %\STATE 
 {\bfseries Parameters:} DP budget $(\varepsilon, \delta)$ \\
 %\STATE
 {\bfseries Output:} Prediction $q$, Counts ($c_0$ and $c_1$) \\
\begin{algorithmic}
 \STATE $c_0 \gets  \mathcal{M}_{LAPLACE}^{\Delta_1}(r, \textbf{\texttt{count\_0}}(X_{rem}, \cdot) , \varepsilon_{node})$\\
 $c_1 \gets  \mathcal{M}_{LAPLACE}^{\Delta_1}(r, \textbf{\texttt{count\_1}}(X_{rem}, \cdot) , \varepsilon_{node})$
 \STATE $q \gets 0$ \textbf{if} $c_0 > c_1$ \textbf{else} 1 
\end{algorithmic}
\end{algorithm}

\textbf{Confidence threshold for minimum support (Line~\ref{line:confidence}).} 
One remaining issue with the proposed smooth sensitivity framework is that the minimum support requirement may jeopardize the DP guarantees. 
For instance, consider $\dataset$ a dataset and a fixed minimum support $\minsupp$, and let $\arule$ a rule. 
Suppose that after applying rule $\arule$, the number of points remaining for classification $\nleft{r}$ is exactly equal to $\minsupp$. 
Let also $\dataset'$ be a dataset neighbouring $\dataset$ that misses one of the samples not caught by $\arule$ in $\dataset$. 
Then, the support of $\dataset'$ after applying rule $r$ is strictly smaller than $\minsupp$ so any rule %$\arule'$ is 
will necessarily be discarded because it is a stopping condition. 
Again, this breaks any DP guarantees, as the resulting model may change significantly due to the absence of a single sample in the dataset. To solve this issue in the proposed algorithm, we consider a threshold  %confidence interval
for minimum support that in most cases preserves the DP guarantees. %of the model. 
Knowing that counting queries have a global sensitivity of $1$, 
%we define the following %heuristic 
%method.
after each split of the dataset, %(\textit{i.e.} each time the rule list is added a rule $r$)    we compute its exact support %of $D$ after $r$. Add 
we add Laplace noise $\sim Lap\big(\frac{\Delta_1 f = 1}{\varepsilon}\big)$ to the noisy support. 
If the noisy support is under a given predefined threshold then we stop here and use the default classification, while otherwise we keep adding rules.
    %\item 
%
%How to determine the threshold? 
To determine the threshold, assume that $\Lambda$ and $\varepsilon$ are fixed and we want a confidence $\mathcal{C} = 0.98$. 
When the added noise is negative (\emph{i.e.}, the noisy support is lower than the exact support), 
%in the worst case 
the algorithm does not add any rule even if the smooth sensitivity computation remains exact. 
However, when the noisy support is above the exact support, we need to assess how large the added noise can be. 
This can be done by studying the distribution of the Laplace noise to determine at what value $t$ it will be above the confidence $\mathcal{C}$. 
%Given the discussion above, 
More precisely, we search for $t>0$ such that :
$
\int_{-\infty}^t Lap(x|b) \, dx \geq \mathcal{C} 
%\iff& \int_{-\infty}^0 \frac{\varepsilon}{2} \exp( x \varepsilon) dx + \int_{0}^t \frac{\varepsilon}{2} \exp(-x \varepsilon) \geq \mathcal{C} \\
%\iff& \Big[\exp(\varepsilon x) \Big]_{-\infty}^0 +\Big[-\exp(-\varepsilon x) \Big]_{0}^t  \geq 2\mathcal{C} \\
%\iff& \exp(-\varepsilon t) \leq - 2\mathcal{C} \\
\iff t \geq - \frac{\log(2) +\log(1- \mathcal{C})}{\varepsilon}$.
The confidence threshold is $\mathcal{T} = 1 + %\coloneqq \Lambda + 
\lfloor t \rfloor$ (Algorithm~\ref{alg:threshold}).

\begin{algorithm}[h!]
\caption{ \textbf{Function \texttt{confidence\_threshold}} : }
\label{alg:threshold}   
 %\STATE 
 {\bfseries Input:} Confidence $\mathcal{C}$ \\
 %\STATE 
 {\bfseries Parameters:} DP budget $(\varepsilon, \delta)$ \\
 %\STATE
 {\bfseries Output:} Threshold  $\mathcal{T}$
\begin{algorithmic}
    \STATE $\mathcal{T} =  \Big\lfloor - \dfrac{\log(2) +\log(1- \mathcal{C})}{\varepsilon_{node}} \Big\rfloor + 1 $
\end{algorithmic}
\end{algorithm}

For instance, with $\varepsilon = 0.1$, and $\mathcal{C} = 0.98$, we obtain $t = \lfloor 6.733 \rfloor + 1 = 7 $.
This means that we can claim with a confidence of $0.98$ that if the algorithm decides to add rules, then it respects the minimal support constraint.  In practice, the confidence $\mathcal{C}$ will only apply to the later rules of the rule list  when the number of samples left becomes scarce. %With a good sampling of a train/test model (in the sense that the samples' distributions in each set are similar), this would lead to very few samples being classified by a ``potentially'' not private rule.
%\jul{ci-dessus: le risk of failure si le support est trop grandi par du bruit positif c'est juste d'ajouter une règle à tort non ? Pas de problème de DP qui elle tient toujours ?}
%\tim{Alors pas exactement : ajouter une règle en pratique ce n'est pas trop grave. Le risque c'est d'ajouter une règle dont on aura calculé un gini pas suffisament bruité car la condition de support minimal, utilisée dans le calcul de la sensibilité lissée, ne serait plus respectée.}
%\seb{ci-dessous préciser peut-être ce que représente a good sampling en pratique}

\textbf{Privacy budget.} Let ($\varepsilon$, $\delta$) be the total privacy budget allocated to the algorithm. 
%\seb{il faudrait être sûr d'avoir expliquer les résultats de compositionalité dans la partie DP avant de pouvoir les utiliser}
Using the sequential and parallel composition for DP mechanisms, we must %first 
determine %what 
the fraction of the privacy budget to allocate per node (\emph{i.e.}, how much privacy budget should be allocated for the choice of each rule). 
We will denote these quantities by $\varepsilon_{node}$ and $\delta_{node}$. Let $K$ the maximum length of a rule list. 
While it is common for tree-based models to display the counts for each leaf (\emph{i.e.}, in our case for each rule), this information should also be made differentially private. 
First in Line~\ref{line:cond-min-sup}, the minimum support condition is verified by applying the Laplace mechanism with global sensitivity (satisfying ($\varepsilon, 0)$-DP). 
Then, the computation of the Gini impurity (Line~\ref{line:gini-smooth-laplace}) is made inside the dataset for each candidate rule and only the rule corresponding to the maximum of these noisy Gini impurity values is returned to the algorithm, which is the \textit{Noisy Max Report} mechanism that only accounts for one DP query (as introduced in Section~\ref{subsec:background:dp}). 
Computing the two noisy counts of the chosen rule (Algorithm~\ref{alg:predDP}) also counts only for one query since the sets of samples caught and not caught are disjoint, which leads to the application of the parallel composition. 
Finally, with sequential composition, it gives us $3$ operations per node, with $2$ achieving pure DP. 
For the default rule, only noisy counts are used and no Gini impurity is computed. 
Therefore, $\varepsilon_{node} = \frac{\varepsilon}{3K - 1}$ %\qquad 
and $\delta_{node} = \frac{\delta}{K-1}$.

\textbf{A variant satisfying pure DP.} While our proposed \DPGRL{} algorithm satisfies approximate ($\varepsilon$, $\delta$)-DP, it is worth observing that the only operation not satisfying pure DP is the noisy max report using the Laplace mechanism along with the smooth sensitivity of the Gini impurity (line~\ref{line:gini-smooth-laplace}). 
This operation satisfies pure DP if the Laplace mechanism is replaced by the Cauchy mechanism within this smooth sensitivity framework, \emph{i.e.,} by replacing line~\ref{line:gini-smooth-laplace} with $ \mathcal{G} \gets \mathcal{M}_{CAUCHY}^{\smooth_{\cdot, \beta}}(r, \mathscr{G}_{X_{rem}}(\cdot), \varepsilon_{node})$.
In such a case, the privacy budget analysis remains unchanged, apart from the $\delta$ parameter which is now $0$, allowing the whole algorithm to yield pure $(\varepsilon,0)$-DP guarantees.
We coin the resulting variant \texttt{sm-Cauchy}. In a nutshell, it is also based on our smooth sensitivity framework, but adds noise from the Cauchy distribution to the computed Gini impurity values to provide pure DP guarantees while still leveraging our smooth sensitivity framework.

%As mentioned previously, most operations are carried out on the side of the dataset. 
%For instance, $X_{rem}$ is only known in its approximate value by the program, but we can still ask the dataset to remove all the samples from $X_{rem}$ caught by $R^{\bigstar}$ without returning us the updated $X_{rem}$. When we want it, we query it using a DP mechanism.
%\mjo{pas très clair pour moi ...}
%\tim{ce que j'essaie de dire, c'est que l'algo n'a jamais accès à tous les gini bruités, seulement à la règle qui minimise les gini. C'est bien un Report Noisy Max et pas une query de tous les ginis.}
\section{Experimental Evaluation}
\label{sec:expe}
%\cmjo{In this section, ...}
In this section, we assess experimentally the effect of smooth sensitivity on the resulting models' accuracy when compared to other approaches based on global sensitivity, for comparable privacy guarantees (pure or approximate DP). We evaluate the performances of the built rule lists in terms of predictive accuracy, robustness to privacy attacks as well as preservation of features' importance.
%\rqe{ajouter et deux variantes de la smooth ?} \ctim{on pourrait dire on the two resulting models' accuracy pour signifier qu'il y en a deux basés sur la smooth}

\subsection{Experimental Settings}
\label{subsec:expe-settings}
%\mjo{implémentation du code, lien vers dépot (footnote), liste des dataset, plateforme d'exécution}
%We assessed the accuracy of our models with real datasets : 
For our experiments, we consider three popular datasets: \namedataset{German Credit}, %\footnote{https://archive.ics.uci.edu/dataset/144/statlog+german+credit+data}, 
\namedataset{Compas} %\footnote{link compas} 
and \namedataset{Adult} %\footnote{link adult} 
in their binarized versions. Sensitive attributes were removed as their use is prohibited to avoid disparate treatment.
The set of rules $\ruleset$ used in these experiments is publicly available. It is made up of %are mined as
conjunctions of up to two Boolean attributes or their negation. 
%as longer rules make the space exploration exponentially more time consuming. 
%\rqe{à modifier : on a dit que $\ruleset$ était connu}
%\ctim{En effet, on pourrait se contenter de dire que les règles sont de 1 ou 2 conjonctions, mais que c'est 'donné' comme ça.}

In \namedataset{German Credit}~\cite{Dua}, the classification task is to predict whether individuals have a good or bad credit score. 
Features are binarized using one-hot encoding for categorical ones and quantiles (2 bins) for numerical ones. 
%Rules are generated as single features with minimum support of 0.25 or conjunctions of two features with minimum support of 0.5. Gender-related features were excluded. 
The resulting dataset contains $1,000$ samples and we consider $49$ premined rules.
For \namedataset{Compas}~\cite{angwin2016machine}, the objective is to predict whether an individual will re-offend within two years or not. 
Features are binarized using one-hot encoding for categorical ones and quantiles (with 5 bins) for numerical ones. 
The resulting dataset contains $6,150$ samples and we have $18$ rules.
The classification task in \namedataset{Adult}~\cite{Dua}
%a larger dataset that gathers records of individuals from the 1994 U.S. census. 
is to predict whether an individual earns more
than $50,000\$$ per year.
%, with gender (male/female) being the sensitive attribute.
Categorical attributes are one-hot encoded and numerical ones are discretized using quantiles (3 bins). 
The resulting dataset contains $48,842$ samples and we use $47$ rules (attributes or their negation). %, with a minimum support of 0.05. 
%\seb{ne pas oublier aussi de compléter les liens et de dire l'objectif de classification pour chaque ensemble de données}
%\jul{je suggèrerais de ne pas dire fair et de juste dire qu'on a enlevé les attributs sensibles dont l'utilisation est prohibée (disparate treatment)}

In our experiments, we build upon the baseline \GRL{} implementation available in the literature~\cite{ferry:hal-04189566}\footnote{https://github.com/ferryjul/ProbabilisticDatasetsReconstruction} and further modify their code to implement our proposed DP mechanisms within the \DPGRL{} and \texttt{sm-Cauchy} algorithms\footnote{The source code will be released publicly upon acceptance.}.
%\jul{Adult et COMPAS: papier satml24, german: papier}
For each value of $\varepsilon$, we average our results over 100 runs with different random seeds to account for train/test distribution (\emph{i.e.}, train/test split of 70/30) and the randomization due to the application of DP. 
%\seb{éventuellement on pourrait préciser si le cadre expérimental correspond à celui utilisé habituellement pour les algos assurant la DP pour les arbres de décision}
The value of $\delta$ was set $\frac{1}{\norm{1}{\dataset}^2}$ %to make the comparison with pure-DP relevant, higher values of $\delta$ could eventually be picked. 
%\jul{Le carré ne devrait pas être en dehors de la norme ?}
and the maximum length for rule lists was set to $K=5$ as we empirically observed that lower values could impede the model accuracy and higher values do not substantially increase accuracy. Importantly, these trends were confirmed by extensive preliminary experiments and were consistent over all methods and datasets, as further discussed in Section~\ref{sub-sec:hyperparameters}.
%\seb{ci-dessus préciser pourquoi des valeurs plus hautes ne sont pas nécessaires (parce que ça n'aide pas l'accuracy ou une autre raison?)}
The other hyperparameters of the proposed algorithm were fixed with preliminary grid search leading to $\mathcal{C} = 0.99$, $\lambda = 0.12$ for \namedataset{German Credit} and $\lambda = 0.05$ for \namedataset{Compas} and  \namedataset{Adult}. 
%
%\mjo{Add $\lambda$ and $\mathcal{C}$. Dire que $K, \lambda, \mathcal{C}$ fixés avec expé préliminaires.}
%
All our experiments are run on an Intel CORE I7-8700 @3.20GHz CPU.
%is available on Github\footnote{Source code of the \DPGRL algorithm will be released publicly.
%\url{https://github.com/UnparalleledSmilingMonster/DP-greedy}

\subsection{Considered Baselines: DP Greedy Rule Lists Algorithms based on global sensitivity}

\label{subsec:best-global-dd}
%\rqe{Ajouter une phrase pour dire quel est le but de cette expé : on oublie la smooth et on fait de la global sensitivité ... et .... }
To assess the effectiveness of our proposed framework leveraging smooth sensitivity, we will compare it with baseline algorithms based on global sensitivity. These baselines are global sensitivity-based variants of our DP greedy rule lists algorithm, and to ensure a fair comparison, the aim of this subsection is to determine their best performing version.
%At each step of Algorithm~\ref{alg:smooth-dp}, the selection of the rule with the best Gini index $R^\bigstar$ (Lines~\ref{line:gini-smooth-laplace-init} and \ref{line:gini-smooth-laplace}) is implemented by Laplace noise with smooth sensitivity. 
%\rqe{pourquoi on revient sur la smooth ? } \ctim{il me semble qu'on rappelait comment on avait fait pour notre algo, puis qu'on explique comment on l'a transformé ensuite pour revenir sur les versions DP globales mais je suis d'accord, il faut reformuler}
%To evaluate the benefit %of the smooth sensitivity 
%compared to the global sensitivity, we
%first determine the best %performing 
%rule %lists model 
%based on the global sensitivity when computing the Gini index. 
%Thus, we implemented
More precisely, in these experiments, we compare two different versions of the greedy rule lists learning algorithm using global sensitivity.
 
\textbf{Noisy Gini.} The first version simply replaces the smooth sensivity of the Gini impurity with its global sensitivity. 
More precisely, in Line~\ref{line:gini-smooth-laplace}, Laplace noise scaled to global sensitivity is added to the Gini Impurity and there is no need to compute the confidence threshold (Line~\ref{line:confidence}) to comply with DP.
Thus, some privacy budget can be saved during that step. 
%\rqe{A relier avec la section III. B où on parle du calcul du Gini et le début de la section III.C où on parle de la global sensitivité du gini (à remettre ici ?)}
%\ctim{TODO, pas eu le temps de le faire. }
%\rqe{Ajouter Noisy Max Report ?}

\textbf{Noisy counts.} The second version leverages the global sensitivity of counting queries (equal to 1) rather than using the global sensitivity of the Gini impurity which is very high (related to the range of possible values for these queries, i.e., $[0,1]$ for Gini impurity values and $[0,\nsamples]$ for counting queries). We first access the counts $\ncapt{r}$ and $\nleft{r}$ for each rule (number of elements caught / not caught by rule $r$) in a differentially private way (through the Laplace mechanism). Using Equations \eqref{eqn:gini_c} and \eqref{eqn:gini_l}, along with the obtained noisy counts, we compute the Gini impurity for each rule, and only keep the rule minimizing it. According to the post-processing property, this quantity remains differentially private. Nonetheless, this access is not a \textit{Noisy Max Report} mechanism anymore but a regular access to all counts for each rule. 
This means that the privacy budget per node needs to be further split for each rule of the ruleset $\ruleset$, which leads to a factor of $1/2|\ruleset|$ in the denominator. 

Note that both considered variants (noisy Gini and noisy counts) satisfy pure DP since they both rely on the Laplace mechanism along with global sensitivity.
%\rqe{AJOUTER $(\varepsilon,0)$-DP algotithms} \newline
%\ctim{Puisqu'on accède avec Laplace c'est effectivement pure DP, mais on pourrait très bien faire avec du gaussien. Est-ce nécessaire de le mettre?}
Alternatively, one could imagine deriving the best Gini impurity using a noisy max report over the counts (i.e., for saving on the privacy budget) but there is no guarantee that the counts returned will belong to the same rule. As such, it becomes impossible to determine the rule with the best Gini using only the noisy max report mechanisms over the counts. 

\begin{figure}[htb]
    \centering
    \includegraphics[width=0.98\linewidth]{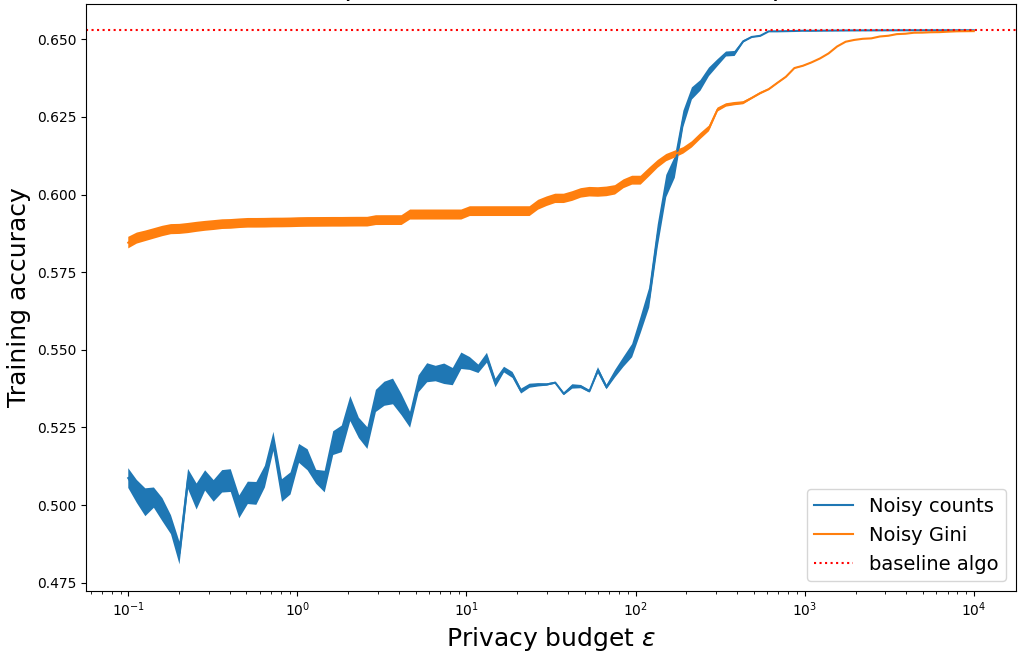}    
    \caption{Comparison of Noisy counts and Noisy Gini versions using global sensitivity (log-scaled), applied on the \namedataset{Compas} dataset.}
    \label{fig:counts-gini-compas}
\end{figure}

%There is no consensus on which value of $\varepsilon$ ~\cite{choice_epsilon} a DP algorithm should bolster. 
In the following experiments, we consider the non-private greedy rule lists algorithm (\GRL) as baseline and the two versions of the pure DP greedy rule lists algorithm based on global sensitivity and Laplace noise (namely, noisy Gini and noisy counts). 
Figure~\ref{fig:counts-gini-compas} displays the training accuracy of these three algorithms when the privacy parameter 
$\varepsilon$ varies from $10^{-1}$ to $10^{+4}$. Note that while such high values do not provide meaningful privacy guarantees (and are not used later in our experiments), they allow verifying the asymptotic behaviour of the two compared variants of global sensitivity-based approaches. In particular, they confirm that both approaches eventually converge towards the non-private variant when $\epsilon$ becomes sufficiently large.
The privacy regime of interest to determine which of the two versions performs the best lies within $\varepsilon \in [0.1, 20]$. Indeed, when $\varepsilon$ goes over $20$, it becomes hard to quantify how the theoretical guarantees apply on realistic settings while a value under $0.1$ leads to poorly performing models. 

Figure~\ref{fig:counts-gini-compas} shows that in the considered privacy regime, a rule list model built using the noisy Gini version (that is, using the global sensitivity of the Gini impurity)
performs better than a model learnt based on noisy counts. However, it is interesting to note that for very large values of the privacy budget $\varepsilon$, the noisy Gini version is slower to reach the accuracy of the baseline model obtained with \GRL{}.
When $\varepsilon$ is high enough, the noise added is so low that the Gini impurity scores are ranked according to their original value hence a consistent result with \GRL{}. 
The model using only the noisy counts remains nonetheless interesting in a setting in which the mined ruleset is pre-processed beforehand to a small  cardinality (\textit{e.g.} less than a hundred). Indeed, the privacy budget is inversely proportional to the cardinal of the ruleset, so better performances can be expected on smaller instances of rulesets.

\begin{figure}[h!]
  \centering
  \begin{subfigure}[t]{0.49\textwidth}
    \centering\includegraphics[width=\textwidth]{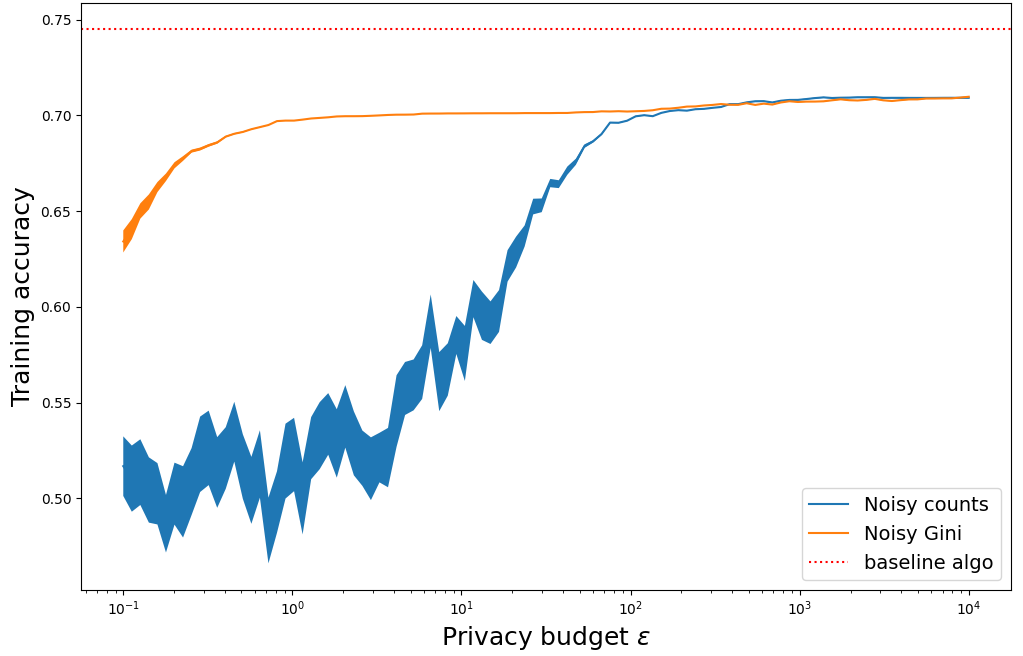}
    \caption{\namedataset{\namedataset{German credit}}}
  \end{subfigure}
  \hfill
 
  \begin{subfigure}[t]{0.49\textwidth}
    \centering\includegraphics[width=\textwidth]{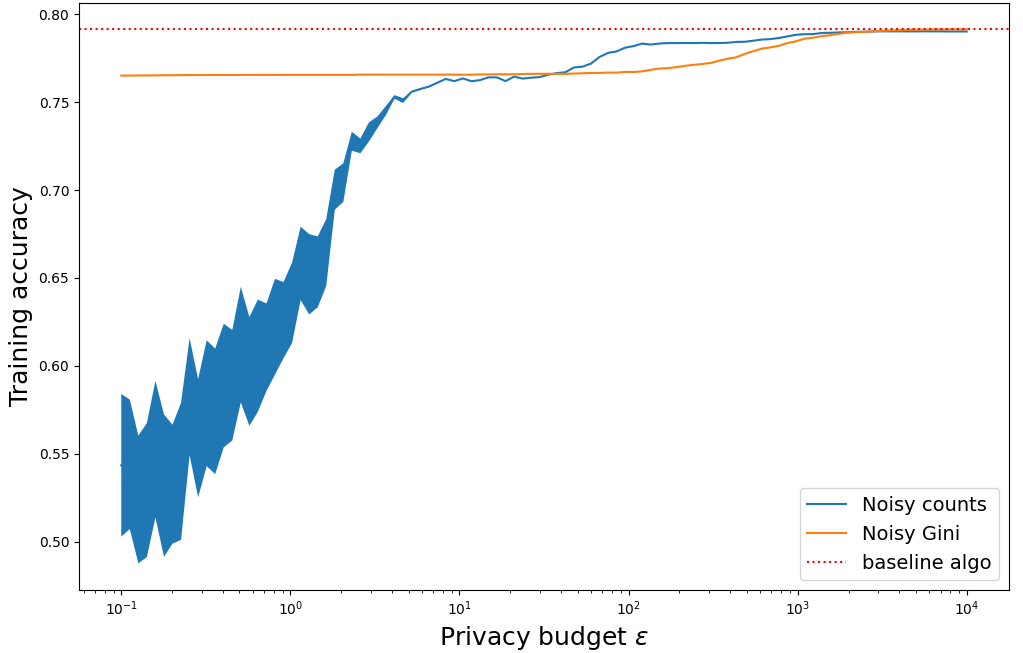}
    \caption{\namedataset{\namedataset{Adult}}}
  \end{subfigure}
  \caption{Comparison of Noisy counts and Noisy Gini versions using global sensitivity (log-scaled) on the \namedataset{German credit} and \namedataset{Adult} datasets.}
\label{fig:acc_gini_counts}
\end{figure}

The same trends are observed in Figure~\ref{fig:acc_gini_counts}
for the \namedataset{German credit} and \namedataset{Adult} datasets when comparing the two versions leveraging global sensitivity to output the best rule. 
The method using the global sensitivity of the Gini impurity (noisy gini) remains the best choice for the considered privacy regimes. Finally, we now focus on this version for the remaining of the experiments. This method will be coined as \texttt{gl-Laplace}, while its variant replacing the Laplace mechanism by the Gaussian mechanism for computing the noisy Gini impurity based on global sensitivity will be coined as \texttt{gl-Gaussian}. Hence, recall that \texttt{gl-Laplace} satisfies pure DP while \texttt{gl-Gaussian} satisfies approximate DP.

%\begin{figure*}[h!]
%  \centering
%  \begin{subfigure}[t]{0.48\textwidth}
   % \centering\includegraphics[width=\textwidth]{counts_vs_gini_german.png}
    %\caption{\namedataset{\namedataset{German credit}}}
  %\end{subfigure}
  %\hfill
  %\begin{subfigure}[t]{0.48\textwidth}
   % \centering\includegraphics[width=\textwidth]{counts_vs_gini_adult.png}
    %\caption{\namedataset{\namedataset{adult}}}
  %\end{subfigure}
  %\caption{Comparison of Noisy counts and Noisy Gini variants  using global sensitivity (log-scaled)}
%\label{fig:acc_gini_counts}
%\end{figure*}

%the noisy Gini version.

% which consists in selecting the rule with the best Gini index $R^\bigstar$ (Lines~\ref{line:gini-smooth-laplace-init} and \ref{line:gini-smooth-laplace}) noised by Laplace noise scaled on global sensitivity (we further discuss this model selection issue in Appendix~\ref{appendix:expe-global-gini}). 

\subsection{Prediction Performance}
\label{subsec:expe-pred-perf}
We now compare the test accuracy of rule list models obtained by Algorithm~\ref{alg:smooth-dp} along with different DP mechanisms.
More precisely, we consider two mechanisms based on smooth sensitivity and either Cauchy (\texttt{sm-Cauchy}) or Laplace (\texttt{sm-Laplace}) noise as well as two mechanisms based on global sensitivity and Gaussian (\texttt{gl-Gaussian}) or Laplace (\texttt{gl-Laplace}) noise.
Finally, we also implemented the Exponential mechanism using the Gini impurity as the utility function for sampling the best rule at each node (\texttt{gl-Exponential}).
For these experiments, we thus consider three pure DP algorithms: \texttt{gl-Laplace}, \texttt{gl-Exponential}, and \texttt{sm-Cauchy} and two approximate DP algorithms: \texttt{gl-Gaussian} and  \texttt{sm-Laplace}. The baseline test accuracy is given by the non-private \GRL{} algorithm.

\begin{figure}
  \centering
  \begin{subfigure}[t]{0.45\textwidth}
    \centering\includegraphics[width=\textwidth]{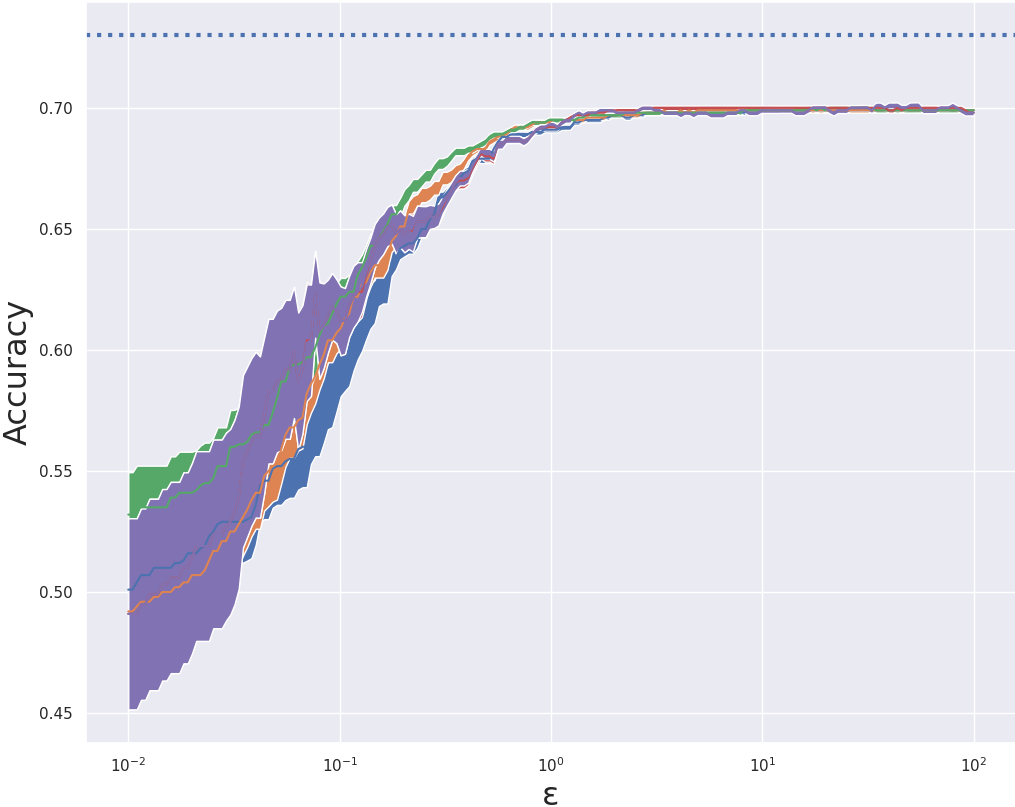}
    \caption{\namedataset{German credit dataset}}
  \end{subfigure}
  \hfill
  \begin{subfigure}[t]{0.45\textwidth}
    \centering\includegraphics[width=\textwidth]{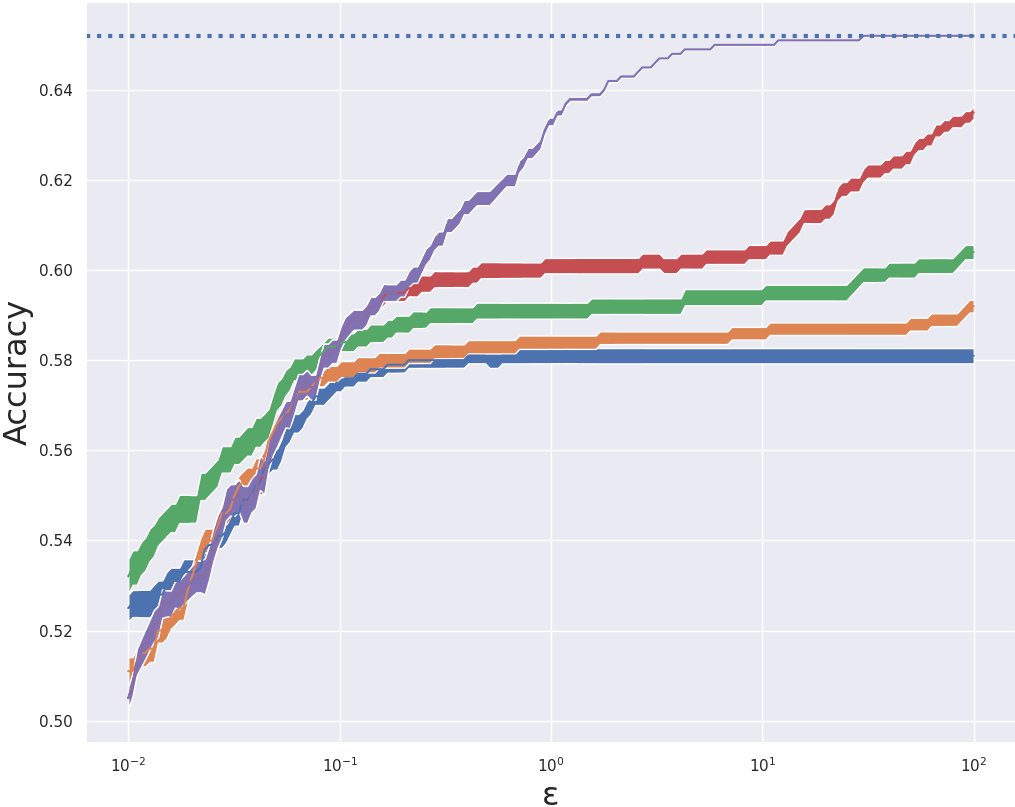}
    \caption{\namedataset{COMPAS dataset}}
  \end{subfigure}
  \hfill
  \begin{subfigure}[t]{0.45\textwidth}
    \centering\includegraphics[width=\textwidth]{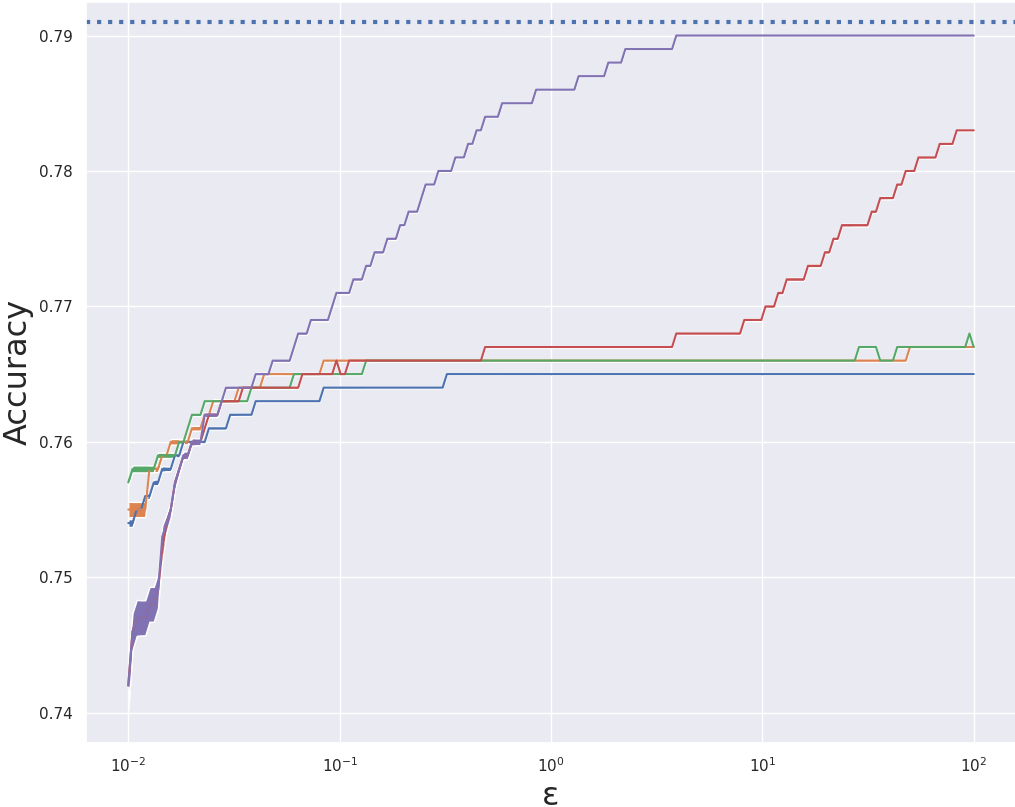}
    \caption{\namedataset{UCI Adult Income dataset}}
  \end{subfigure}  
  
  \vspace{0.5em} % Adjust the vertical space between the rows
\begin{subfigure}{0.47\textwidth}
    \centering
    \includegraphics[width=\linewidth]{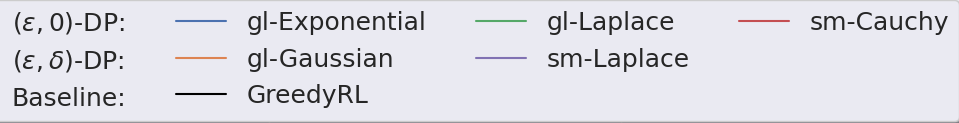}
\end{subfigure}
  \vspace{0.5em} % Adjust the vertical space between the rows
  \caption{Comparison based on the test accuracy of different DP rule list algorithms.}
\label{fig:accuracy}
\end{figure}

In our experiments, we consider privacy budgets $\varepsilon \in [0.01, 100]$ for a total of 200 values with $\varepsilon$ uniformly distributed across the logarithmic scale. 
The results, averaged over $100$ runs as described in Section~\ref{subsec:expe-settings}, are displayed in Figure~\ref{fig:accuracy} and the test accuracy for $\varepsilon=10$ is reported in the right part of Table~\ref{table:vulnerability}. Note that we use a logarithmic scale for the values of $\varepsilon$ on the x-axis of Figure~\ref{fig:accuracy} as is often the case in the literature~\cite{DPSGD}, in order to avoid hiding the tightest privacy budgets (\emph{e.g.,} $\varepsilon \leq 1$).

As shown in Figure~\ref{fig:accuracy}, the two variants based on the smooth sensitivity framework 
perform particularly well for relatively large datasets (\namedataset{Compas} and \namedataset{Adult}). In addition, for \namedataset{Compas} and \namedataset{Adult}, the convergence of the approaches based on smooth sensitivity to the baseline model is very steep. In contrast, %whereas 
DP mechanisms based on global sensitivity usually converge around $\varepsilon \approx 10^3$. 
For $\varepsilon \geq 0.1$, the mechanisms based on smooth sensitivity either match or outperform the standard global DP approaches.
Importantly, the \texttt{sm-Laplace} mechanism consistently and largely outperforms all other approaches on a wide range of privacy budgets of interest. Focusing on pure DP methods, the \texttt{sm-Cauchy} mechanism consistently performs better than the global sensitivity based frameworks. These experiments confirm the theoretical analysis: for both approximate and pure DP, and for a wide range of privacy budgets, the use of smooth sensitivity in place of global sensitivity allows for better accuracy-privacy trade-offs.

We now compare the two variants based on smooth sensitivity, namely \texttt{sm-Cauchy}, satisfying pure DP, and \texttt{sm-Laplace}, providing approximate DP guarantees. The \texttt{sm-Cauchy} accuracy is much slower to converge than the latter, as can be observed on the \namedataset{Compas} and \namedataset{Adult} datasets.
The Cauchy distribution has a polynomial decaying tail, which is much heavier than the exponential decaying tail of the Laplace distribution. 
Thus, out of the many random noise values generated at each step of the algorithm, a few might end up far from the average amplitude, which might deteriorate significantly the accuracy. %utility. 
%\seb{attention, on parle du mécanisme de Cauchy ci-dessous mais même si on avait donné sa formule je pense pas qu'on l'avait nommé précédemment}
As a consequence, although the \texttt{sm-Cauchy} mechanism provides a good alternative to pure DP mechanisms based on the global sensitivity, we advise to replace it by its Laplace counterpart even if the provided privacy guarantees are slightly weaker. 

Compared to the differentially private random forest proposed in~\cite{dpsmooth-forest}, we incur at $\varepsilon=1$ a significantly lower accuracy loss with respect to the non-private model. For this level of privacy, our smooth sensitivity-based \texttt{sm-Laplace} algorithm has less than 0.5 absolute accuracy decrease (the accuracy of the proposed DP algorithm is 78.7\% vs 79.1\% for the non private version) against at least 1.0 for theirs (the accuracy of the DP algorithm is 82\% and the non DP version is about 83\%).

% the non-private model and the private model based on the smooth sensitivity.}

%\cmjo{Remark that} while the $\GRL{}$ has subpar performances on \namedataset{Adult}, it reaches \textit{state of the art} accuracy on \namedataset{Compas}.

%\rqe{légende à changer de Figure\ref{fig:accuracy}}

We now focus on the three best performing methods, namely our proposed smooth sensitivity based \texttt{sm-Laplace} and \texttt{sm-Cauchy} mechanisms as well as the global sensitivity baseline \texttt{gl-Laplace}. We compute and display the standard error of the empirical mean estimator $\theta$ whose formula is as follows: 
$$ err(\theta) = \sqrt{\frac{\text{Var}(\theta)}{\Gamma}}$$ 
where $\Gamma$ is the number of observations for a given set of hyperparameters.

\begin{table}[h!]
  \centering
\caption{Standard error of the empirical mean estimator across datasets for global and smooth sensitivity based DP mechanisms with respect to test set accuracy.}
\label{fig:variances}
\begin{subfigure}[b]{0.45\textwidth}  \centering
\caption{Standard error for the \texttt{sm-Laplace} algorithm.}
\begin{tabular}{@{}cccc@{}}
\toprule
$\varepsilon$ & \namedataset{German} & \namedataset{Compas} & \namedataset{Adult} \\ \midrule
0.01          & 2.0e-02              & 5.0e-03              & 4.2e-03             \\
0.1           & 1.2e-02              & 5.0e-03              & 1.0e-03             \\
1             & 3.6e-03              & 2.9e-03              & 4.7e-04             \\
10            & 3.3e-03              & 1.2e-03              & 4.3e-04             \\
100           & 3.4e-03              & 1.1e-03              & 4.1e-04             \\ \bottomrule
\end{tabular}
\end{subfigure}

\vspace{5pt}

\begin{subfigure}[b]{0.45\textwidth}  \centering
\caption{Standard error for the \texttt{sm-Cauchy} algorithm.}
\begin{tabular}{@{}cccc@{}}
\toprule
$\varepsilon$ & \namedataset{German} & \namedataset{Compas} & \namedataset{Adult} \\ \midrule
0.01          & 2.0e-02              & 5.0e-03              & 4.2e-03             \\
0.1           & 1.2e-02              & 5.0e-03              & 8.4e-04             \\
1             & 3.6e-03              & 4.1e-03              & 8.5e-04             \\
10            & 3.1e-03              & 4.0e-03              & 9.9e-04             \\
100           & 3.1e-03              & 3.6e-03              & 8.0e-04             \\ \bottomrule
\end{tabular}
\end{subfigure}

 \vspace{5pt} 
\begin{subfigure}[b]{0.45\textwidth}  \centering
\caption{Standard error for the \texttt{gl-Laplace} algorithm.}
\begin{tabular}{@{}cccc@{}}
\toprule
$\varepsilon$ & \namedataset{German} & \namedataset{Compas} & \namedataset{Adult} \\ \midrule
0.01          & 1.3e-02              & 6.2e-03              & 1.7e-03             \\
0.1           & 8.8e-03              & 4.5e-03              & 8.0e-04             \\
1             & 3.3e-03              & 4.2e-03              & 8.0e-04             \\
10            & 3.1e-03              & 4.2e-03              & 8.0e-04             \\
100           & 3.0e-03              & 4.0e-03              & 8.9e-04             \\ \bottomrule
\end{tabular}
\end{subfigure} 
\end{table}

Table~\ref{fig:variances} presents the standard error of the empirical mean estimator with respect to the random seeds at given values of $\varepsilon$. 
We observe a high standard error on  accuracy at low $\varepsilon$ for the smooth sensitivity based models (under 1e-2, standard error can be considered low because it means the accuracy hovers by less than $\pm 0.01$). 
Global sensitivity models yield a similar standard error at very low $\varepsilon$. 
As $\varepsilon$ values goes up, the standard errors decreases which means that the models' behaviour becomes more deterministic. 
We also observe that the error decreases when the size of the dataset is higher, although by a small margin between 
$\namedataset{Compas}$ and $\namedataset{Adult}$ but the standard error is five times higher at $\varepsilon=0.01$ on the \namedataset{german-credit} dataset. 
Overall, the smooth sensitivity model using Laplace noise consistently has an equivalent or lower standard error than the global sensitivity model for $\varepsilon \geq 1$. 
The discrepancy observed for the error of the Cauchy noise model is most likely based on the previous observation about Cauchy distribution's wide tail. 
\newline
We can provide an explanation for this deviation compared to global sensitivity models, and it will partly answer why the smooth sensitivity models have poorer performances at low $\varepsilon$. 
Indeed, the confidence threshold becomes exceedingly high for these privacy values and the minimum support condition is therefore more likely to fail, which causes the model to output only one rule. Naturally, in that case of underfitting, the smooth sensitivity models cannot perform as well as the classic DP models. This issue could eventually be tackled by assigning more privacy budget to the minimum support condition and less to the noisy Gini impurity computation. This asymptotic behavior however disappears quickly, especially for larger datasets because the confidence threshold variable is independent of the actual value of the minimum support and its value gets relatively smaller as the size of the dataset increases.

\subsection{Hyperparameters and Fine-Tuning}
\label{sub-sec:hyperparameters}

We now provide some insights regarding the influence of the hyperparameters of Algorithm~\ref{alg:smooth-dp} on the accuracy of the resulting rule list models. 
Consequently, we explain  how these hyperparameters' values were chosen. 

Naturally, with rules of higher cardinality (i.e., a higher number of conditions on the attributes), we can expect higher accuracy since the splits would be more refined. 
Observe that it would not affect the privacy budget of the model since the \textit{Noisy Max Report} 
is independent of the number of elements from which the \texttt{argmax} is searched. 
However, it yields an exponential increase in time complexity.  

%Max length of a rule list K 
Optimizing the maximum number $K$ of rules in the rule list proves to be interesting. 
Indeed, the maximum number of rules $K$ heavily influences the privacy budget per node, but it is also dependent on the minimum support condition $\lambda$. 
Namely, there can be no more than $\min(K, \lfloor \frac{1}{\lambda} \rfloor )$ rules in the output rule list. 
Decreasing $\lambda$ enables the inclusion of more rules, but there is a trade-off with the precision of the Laplace noise using smooth sensitivity (the higher $\lambda$  the less noise added). 
Overall, our smooth sensitivity method consistently beats the global sensitivity methods for any value of $K$. 
%(naturally without considering extreme values such as $K < 3$ or $\lambda \geq 0.25$ since it is simply a matter of model underfitting). 
A value of $K=5$ was on average the best performing for all models. For $K=7$ this value was most of the time not reached as the minimum support condition was not achieved anymore. Models using global sensitivity were also terminated before reaching this depth since the algorithm stops when the Gini is not improved anymore.

%\rqe{est-ce qu'il y a qque chose à dire sur  Confidence $\mathcal{C}$ ?}
%\ctim{Plus tu la fixes haute, plus l'algo est susceptible de s'arrêter tôt.}

\subsection{Robustness to Privacy Attacks}
\label{subsec:expe-attacks}

The protection provided by DP aims at hiding the contribution of any individual example to the output of a computation.
Then, it is natural to evaluate it in practice using Membership Inference Attacks (MIAs)~\cite{membership-inference}, whose objective is to determine whether an individual was part of a given model's training set or not. 
Indeed, performing such attacks on both the original greedy rule lists and their DP counterparts, and comparing the MIA success rate, empirically quantifies the effectiveness of the DP protection. 
However, this approach has two main drawbacks. 
First, one has to select which MIA(s) to run, and different attacks can come with different success rates. 
Second, we implemented and used several popular attacks from the literature, and they struggled attacking even the original (non-DP) model, as reported in the Appendix~\ref{subsec:mia-results}. 
An intuitive explanation lies in the simplicity of our considered models: while the output of a deep neural network is a numerical value which can virtually take any value, a rule list classifies an example using one of $K$ rules in which $K$ is reasonably small. While this constitutes an important argument in favor of the use of rule list models, it also makes the empirical assessment of DP more difficult.
Thus, for our empirical analysis of the rule lists' robustness to privacy attacks,
we rather leverage the (model-agnostic) notion of \emph{distributional overfitting} of a model, introduced by~\cite{vulnerability}. 
In a nutshell, \emph{distributional overfitting} aims at quantifying how the model output distribution varies between samples inside and outside the training set. 
It is thus highly correlated to the vulnerability of a model to MIAs, and can be seen as an upper-bound over their success.
Since rule lists are interpretable models, it is entirely possible to know what rule caught a given sample solely by iterating through the successive rules until one evaluates to true for the designated sample. 
For this reason, we have slightly modified the formula for distributional overfit computation to account for the knowledge that the adversary could get from the structure of the model.
More precisely, we define the distributional-overfitting distance with respect to label $y$ as:

\[\tau(y) = \dfrac{1}{2}\sum_{r \in RL} \Big|\p[r | y, M = 1] - \p[r | y, M = 0] \Big|\]
in which $\p[r | y, M]$ is the probability that a sample with label $y$ (from the training set ($M=1$) or outside ($M=0$)) is captured by rule $r \in RL$. 

The overall vulnerability of a model introduced in~\cite{vulnerability} is then computed as the average of distributional-overfitting distances:
\[ V = \frac{1}{2} + \frac{1}{2} \sum_{y \in \{0,1\}} \p[y] \times \tau(y)\]
Intuitively, when measured on finite training and test sets, it measures how much the proportions of samples from each possible label differ among the different rules.  
If the model's outputs have the exact same distributions inside and outside the training set, the vulnerability is $0.5$ indicating that the expected success of a MIA is that of a random guess. 
We report in Table~\ref{table:vulnerability} the overall vulnerabilities measured on rule lists built with or without the use of DP. %within the greedy learning algorithm. 
Consistent with our preliminary observations that the greedily-built rule lists are resilient to MIAs, the vulnerabilities of both the DP and non-DP models are very low. Nevertheless, we observe that non-DP models consistently exhibit slightly higher vulnerability values than their DP counterparts. This is particularly the case for the smallest dataset, namely \namedataset{German Credit}, which highlights the relevance of a DP protection for such low-data regime.
%We believe this methodology remains nonetheless quite thorough and relevant to apply for models whose vulnerability for the baseline is higher. 
% Due to its indep of particular impl
%Table~\ref{table:vulnerability} presents the overall vulnerability on the same datasets. 
%Due to the fact that the greedy rule list does not yield the best accuracy compared to sophisticated models (\emph{e.g.}, CART), this leads to a low vulnerability for the vanilla model. 
%\mjo{C'est bizarre de terminer en faisant ref à l'annexe. Donner une phrase de synthèse de ce qu'il y a dans l'annexe (penser à la numéroter)}

\begin{table}[] %h!
\caption{Test accuracy and overall vulnerability of the greedy rule lists algorithm and its DP counterpart over 100 runs. Notation $0.507^+$ indicates that the non truncated value was greater than the displayed one, and $0.507^-$ indicates it was smaller.}
\label{table:vulnerability}
\begin{center}
\begin{scriptsize}
%\begin{sc}
\begin{tabular}{lcccr}
\toprule
Dataset & Method & Vulnerability & Accuracy \\
\midrule
\namedataset{Compas} & \GRL  &  $0.507^+ \pm \num{4e-6}$ & $0.660 \pm \num{8e-5}$\\ %
\namedataset{Compas} & \DPGRL  & $0.507^- \pm \num{4e-6}$ & 0.658 $\pm \num{1e-4}$ \\
\namedataset{German}   & \GRL  & $0.524 \pm \num{3e-5}$ &  0.711 $\pm \num{5e-4} $\\
\namedataset{German}   & \DPGRL  & $0.516 \pm \num{5e-5}$  & 0.683 $\pm\num{1e-3} $ \\
\namedataset{Adult} & \GRL  & $0.502^+ \pm \num{7e-7}$ & 0.798 $\pm\num{1e-5} $ \\ 
\namedataset{Adult} & \DPGRL  & $0.502^- \pm \num{6e-7}$  & 0.795 $\pm\num{1e-5} $  \\
\bottomrule
\end{tabular}
%\end{sc}
\end{scriptsize}
\end{center}
\end{table}

\subsection{Preservation of Feature Importance}
\label{subsec:interpretability}

In order to assess the effect of DP on feature importance, we use the methodology proposed by Dai et al.~\cite{top-k-feature}:
\begin{enumerate}
    \item Consider a reference model $\rlist_{ref}$ trained with the \GRL{} baseline and a DP model $\rlist_{DP}$ obtained using one of our proposed algorithms. For each of them, compute their top-k-features using Feature Permutation Importance~\cite{breiman-rf}, which calculates how much a feature is correlated to the output of the model. We denote the resulting sets $\text{top}(k,\rlist_{ref})$ and $\text{top}(k,\rlist_{DP})$ 
    \item Compute their intersection ratio:\\ $I_k = \dfrac{\big|\text{top}(k,\rlist_{ref}) \cap \text{top}(k,\rlist_{DP})\big|}{k}$
\end{enumerate}
Higher values for $I_k$ indicate a smaller distortion of the feature importance values, hence a better preservation of this property despite the application of DP.
We focus on the \namedataset{Adult} dataset, considering the top-$k$ features for $k=7$ and maximum length of $K=5$ for the learnt rule lists.
We evaluate how the features selected by \GRL{} are conserved when the rule lists are built by \DPGRL{} and \texttt{gl-Laplace}. Intuitively, our objective is to assess if the noise added to comply with DP significantly distorts the most influential features, and whether this trend is different for smooth sensitivity and global sensitivity based frameworks. 
 
We report our results in Table~\ref{table:feature-importance} for two different privacy budgets, namely $\varepsilon=1$ and $\varepsilon=10$. For both considered values of $\varepsilon$, the \texttt{gl-Laplace} model has poor feature intersection ratio, not exceeding 0.3. In comparison, the smooth sensitivity based \DPGRL{} has a feature intersection ratio of at least 0.5 and we observe that as $\varepsilon$ goes up (\textit{i.e.} when the privacy guarantees are lower), the feature intersection ratio also increases to reach nearly 0.7 for $\varepsilon=10$. The results consistently show that the smooth sensitivity based \DPGRL{} yields higher feature intersection than the global sensitivity based \texttt{gl-Laplace} in both cases and we infer that this model better conserves feature importance values, less distorting explainability.

\begin{table}[h!]
\caption{Feature Importance Analysis (\GRL{} being the baseline method) over 100 runs.}
\label{table:feature-importance}
\begin{center}
\begin{small}
\begin{tabular}{lcc}
\toprule
$\varepsilon$ &Method & Feature Intersection Ratio \\
\midrule
1 & \DPGRL  & $0.542 \pm 0.03$ \\
1 & \texttt{gl-Laplace} &$0.316 \pm 0.04 $ \\
10 & \DPGRL  & $0.684 \pm 0.03$ \\
10 & \texttt{gl-Laplace}  &  $0.308 \pm 0.04$ \\
\bottomrule
\end{tabular}
\end{small}
\end{center}
\end{table}

Since the noise added by the smooth sensitivity model is much lower than using global sensitivity, the results are expected as the same rules tend to be selected for \GRL{} and the smooth \DPGRL{}. Overall, the feature intersection ratio follows a tendency of increasing when $\varepsilon$ goes up. %For the global sensitivity model, we do not observe this for $\varepsilon = 1 \to 10$, this is not an issue as we remind the reader that \GRL{} is not certifiably optimal itself.
\section{Discussion}
\label{sec:concl}
%\cjul{In this paper, we proposed ... Our experiments illustrate that... Insightful research avenues include...}
In this paper, we have proposed a new mechanism for learning interpretable models with DP guarantees, leveraging the smooth sensitivity of the Gini impurity. 
This work directly addresses a key challenge pointed out in the literature~\cite{decision-tree}. 
Our experiments illustrated that this new method, with equivalent privacy guarantees, offers a considerable reduction of the accuracy loss compared to the differentially-private methods using global sensitivity. 

Several promising research directions emerge from this study. 
First, adaptive composition~\cite{foundations_DP} could be leveraged to tighten the computation of the privacy budget of our proposed algorithms. 
Second, it would be insightful to integrate our closed formula for the smooth sensitivity of the Gini impurity within different mechanisms. For instance, the inverse sensitivity mechanism~\cite{asi2020instance} consists in an Exponential Mechanism scaled with the inverse sensitivity (or path length): rather than measuring the variations of a function between two adjacent databases, the authors compute the minimum distance from a database to reach another one achieving a chosen target value. While the exact path length introduced in their paper is often intractable, they derived a method using smooth sensitivity to approximate the path length. This method provides a better alternative to classic noisy mechanisms using smooth sensitivity for pure-DP mechanisms since they do not have to use heavy-tailed distributions such as Cauchy. 
Third, the smooth sensitivity of the Gini impurity could be used to train DP decision trees, random forests, or other types of interpretable models. As was the case for rule lists, one can expect an improvement of the resulting accuracy-privacy trade-offs. Furthermore, the supports of the leaves of a decision tree are disjoint, which is not necessarily the case for the rules within a rule list. Then, parallel composition can be better leveraged, resulting in tighter privacy guarantees than for rule list models. This could lead to even greater improvements of the accuracy-privacy trade-offs. 
Finally, integrating DP within certifiably optimal learning algorithms such as \texttt{CORELS} is another promising research avenue. Indeed, this tree-based algorithm could naturally be leveraged to implement the exponential mechanism. However, several technical aspects should be carefully considered. In particular, \texttt{CORELS} relies on optimality-based bounds to efficiently prune out solutions, but these bounds impede DP. For instance, for all permutations of a given set of rules, \texttt{CORELS} only considers the permutation yielding to the best accuracy for further exploration of the space, which breaks the DP guarantees as the probability of outputting a sub-optimal rule list becomes exactly $0$. A possible approach to address this is to deactivate all bounds but then \texttt{CORELS} essentially breaks down to a complete exploration of the search space, which highly impacts its performance.

\bibliographystyle{IEEEtran}
\bibliography{IEEEabrv,references}

\onecolumn 

\appendices

\section{Key Results for Differential Privacy from the Literature}

\subsection{Distance between Databases}
\label{appendix:dist-db}
We consider tabular datasets, with features being $0-1$ encoded and binary labels.
In particular, we assume that a sample from the dataset is made of $\nfeatures$ features and one label. The universe $\universe$ of all possible samples is therefore finite, with cardinality $2^{\nfeatures+1}$. 
An element $a$ of $\universe$ can be expanded to its tuple form as $(a_1, \dots, a_\nfeatures, a_{\nfeatures+1})$ in which $a_{\nfeatures+1}$ is the label. 
We define the order relation $\preceq$ on $\universe$ such that for $(a,b) \in \universe$,
\begin{equation*}
a \preceq b \iff
    \begin{cases}
    \exists i \in \llbracket 1, \nfeatures +1 \rrbracket, \forall k \in \llbracket 1, i -1 \rrbracket, \ a_k \leq b_k \text{ and } a_i < b_i &\\
    or \\
\forall i \in \llbracket 1, \nfeatures +1 \rrbracket, a_i = b_i&
    \end{cases}           
\end{equation*}  

$\preceq$ yields the symmetric, reflexive and transitive properties and all elements can be compared within $\universe$ so this is a total order relation. 
As such, $(\universe, \preceq)$ is a totally ordered set. 
We can now introduce the expanded notation for datasets. 
A dataset $x$ is a collection of elements of $\universe$ that we write as a tuple $x = (x_0,\dots, x_{|\universe|}) \in \mathbb{N}^{|\universe|}$ such that $x_i$ denotes the number of elements of $\universe$ of type $i$ stored in the database $x$. 
The number of elements in a dataset $x$ is given by the formula:
$\norm{1}{x}\coloneqq \sum_{i=0}^{|\universe|} x_i $. %\newline
 
With this notation, it is easy to interpret the notion of distances between dataset as the $L1$-norm of their difference. We say that two datasets $x,y$ are adjacent if they vary only by one element (\textit{i.e.} $||x-y ||_1 = 1$). 

\subsection{Composition and Post-Processing Properties}
\label{appendix:comp_post_dp}
%The DP-mechanisms presented above possess nice properties to use them in conjunction. DP would not yield any relevance were the entity using the privatized data able to untangle it. 
The \textit{Post-processing theorem} guarantees that one cannot make a differentially private algorithm less private due to post-processing  (unless this post-processing itself accesses the data).
\begin{theorem}[Post-processing theorem]
Let $\algo : \mathbb{N}^{|\universe|} \rightarrow \mathscr{Y}$ be an $(\varepsilon, \delta)$- differentially private algorithm. For any function $f:\mathscr{Y} \rightarrow \mathscr{Z}$, the composition $f \circ \algo :\mathbb{N}^{|\universe|} \rightarrow \mathscr{Z} $ is $(\varepsilon, \delta)$-DP. 
\end{theorem}
%\mjo{le terme DP randomized algo n'a pas été introduit ?}
%\mjo{pas des raisons de place, je mettrai la preuve en annexe (mais garde là pour ton rapport à KTH)}

The differentially private mechanisms considered in this paper all apply on a $\mathbb{R}^k$ valued function. 
The \textit{composition of differentially private mechanisms} enables us to scale up from functions to algorithms. 
More precisely, composition tends to deteriorate the privacy guarantees but to a measurable extent. 
%It all depends on how the composition is applied. 
\textit{Sequential composition} occurs when when several differentially private mechanisms, denoted $\amechanind{1}, \dots, \amechanind{p}$ with respective DP-coefficients $(\varepsilon_1, \dots, \varepsilon_p)$ are applied onto the same dataset $x$. Then the generated output : $(\amechanind{1}(x), \dots, \amechanind{p}(x))$ satisfies $(\sum_{i=1}^p \varepsilon_1)$-DP. 
For \textit{parallel composition}, the differentially private mechanisms denoted $\amechanind{1}, \dots, \amechanind{p}$ are applied into disjoints subsets of a given dataset $x = \amalg_{i=1}^p x_i$ then the generated output : $(\amechanind{1}(x), \dots, \amechanind{p}(x))$ satisfies $(\max_{i=1}^p \varepsilon_i)$-DP.  

%\newpage

\subsection{Proof of the iterative Computation Lemma of Smooth Sensitivity (Lemma~\ref{lemma-smooth}, from~\cite{nissim})}
\label{appendix:lemma-smooth-iterative-computation}
Let $\dataset$ and $\dataset'$ denote two datasets. 
Note that since : $\{\dataset' \in \mathbb{N}^{|\universe|} \ :  \norm{1}{\dataset'-\dataset} \leq k \} \subset \{\dataset' \in \mathbb{N}^{|\universe|} \ :  \norm{1}{\dataset'-\dataset} \leq k+1 \}$ we have that $\forall k \in \mathbb{N}, \mathcal{T}_{k+1}(\dataset) \geq \mathcal{T}_k(\dataset)$. 
\begin{align*}
    \smooth_{f,\beta}(\dataset)&=  \max_{\dataset' \in \mathbb{N}^{|\universe|}} \ls_f(\dataset') e^{-\beta ||\dataset-\dataset'||_1} \\
    &=  \max_{k \in \{0,\dots, n\}} \max_{\substack{\dataset' \in \mathbb{N}^{|\universe|}\\ ||\dataset - \dataset'||_1 = k}} \ls_f(\dataset') e^{-\beta ||\dataset-\dataset'||_1}\\
    &=\max_{k \in \{0,\dots, n\}}e^{-\beta k} \max_{\substack{\dataset' \in \mathbb{N}^{|\universe|}\\ ||\dataset - \dataset'||_1 = k}} \ls_f(\dataset') \\
        &=\max_{k \in \{0,\dots, n\}}e^{-\beta k} \mathcal{T}_k(\dataset) \\
\end{align*}
The transition from the penultimate to the final line is tricky. $\mathcal{T}_k(\dataset)$ is a max over the closed ball of elements at distance at most $k$ of $\dataset$, not the sphere of elements at distance $k$ exactly from $\dataset$. Note that since we consider datasets, the distance can only be an integer.

\begin{align*}
    \mathcal{T}_{k+1}(\dataset) &= \max( \max_{\substack{\dataset' \in \mathbb{N}^{|\universe|}\\ ||\dataset' - \dataset||_1 < k+1}} \ls_f(\dataset'), \max_{\substack{\dataset' \in \mathbb{N}^{|\universe|}\\ ||\dataset' - \dataset||_1 = k+1}} \ls_f(\dataset') ) \\
    &= \max( \max_{\substack{\dataset' \in \mathbb{N}^{|\universe|}\\ ||\dataset' - \dataset||_1 \leq k}} \ls_f(\dataset'), \max_{\substack{\dataset' \in \mathbb{N}^{|\universe|}\\ ||\dataset' - \dataset||_1 = k+1}} \ls_f(\dataset') ) \\
     &= \max(  \mathcal{T}_k(\dataset),  \max_{\substack{\dataset' \in \mathbb{N}^{|\universe|}\\ ||\dataset' - \dataset||_1 = k+1}} \ls_f(\dataset') ) \\
\end{align*}

Since $\beta >0$, $e^{-\beta k} > e^{-\beta(k+1) }$ therefore $e^{-\beta k}  \mathcal{T}_k(\dataset) >  \mathcal{T}_k(\dataset)  e^{-\beta(k+1)}$. 
However, the quantity $e^{-\beta k}  \mathcal{T}_k(\dataset)$ appears in the computation of $\smooth_{f,\beta}(\dataset)$ and since it is strictly greater than the left term of $\mathcal{T}_{k+1}(\dataset)$ we can ignore this term and it is equivalent to compute $ \ls_f(\dataset')$ either on the ball or on the sphere of radius $k$ in that case.

\section{Proof of the Smooth Sensitivity of the Gini Impurity (Theorem~\ref{theorem-smooth})}
\label{appendix:proof}

\subsection{Case 1: For a minimum support of $1$}
\label{appendix:smooth-sensitivity-1}
Assume first that the minimum support $\Lambda$ is 1. 
We will then generalize the result.
To match with the notations used so far, we will consider a dataset $x \in \databases$ and suppose we take interest at the first node splitting this dataset (it is only a matter of notation), we can therefore rewrite the local sensitivity of the Gini impurity at $x$ as:
        \[ LS_\mathcal{G}(x) = 1 - \Big( \dfrac{\norm{1}{x}}{\norm{1}{x}+1} \Big)^2 - \Big( \dfrac{1}{\norm{1}{x}+1} \Big)^2\]

    Consider the function
\[g: \left|   \begin{array}{ccl}
    \mathbb{R}^+ & \longrightarrow  & [0,1] \\
    x & \longmapsto & 1 - \Big( \dfrac{x}{x+1} \Big)^2 - \Big( \dfrac{1}{x+1} \Big)^2 \\
  \end{array} \right. \]

Note that : $LS_\mathcal{G} \equiv g \ \circ \ \norm{1}{\cdot}$. $g$ is differentiable on $\mathbb{R}^+$ and $\forall x \in \mathbb{R}^+,    g'(x) = \dfrac{2(1-x)}{(x+1)^3}$

%\begin{adjustbox}{width=\linewidth}
\begin{tikzpicture}
   \tkzTabInit{$x$ / 1 , $g'(x)$ / 1, $g$ /2}{$0$, $1$,$+\infty$}
   \tkzTabLine{ ,+, z, -, }
   \tkzTabVar{-/ $0$, +/ $\frac{1}{2}$, -/ $0$}
\end{tikzpicture}
%\end{adjustbox}

As a reminder, we are trying to determine the smooth sensitivity of the Gini impurity: 

\[ \smooth_{ \mathcal{G}, \beta}(x) = \max_{k \in \mathbb{N}} \ e^{-k \beta} \mathcal{T}_k(x) \]

where \[
 \mathcal{T}_k(x) = \max_{\substack{y \in \mathbb{N}^{|\universe|}\\ ||y - x||_1 \leq k}} LS_\mathcal{G}(y)  = \max_{\substack{y \in \mathbb{N}^{|\universe|}\\ ||y - x||_1 \leq k}} g \ \circ \ \norm{1}{y} = \max_{\substack{y \in
 \mathbb{N}  \\ y \in [\norm{1}{x}-k, \norm{1}{x}+k]} } g(y)
\]
We consider that $\norm{1}{x} \geq 1$ as we do not build nodes when there are no samples to classify.
$ [\norm{1}{x}-k, \norm{1}{x}+k]$ is an interval with integer bounds. 
With the previous study of $g$ monotonicity, this maximum is reached in $y= \max (1, \norm{1}{x}-k)$. %\newline

\textbf{Explanation:} \begin{itemize}
    \item if $k \geq \norm{1}{x} \geq 1 $ then $1 \in [\norm{1}{x}-k, \norm{1}{x}+k]$ so the maximum is the global maximum of $g$ : $1=\max (1, \norm{1}{x}-k)$.
    \item if $k < \norm{1}{x}$ then $[\norm{1}{x}-k, \norm{1}{x}+k] \subset [1, + \infty]$  and $g$ is monotonously decreasing on $[1, + \infty[$ so the maximum is the leftmost bound of the interval : $\norm{1}{x}-k = \max (1, \norm{1}{x}-k)$.
\end{itemize}
$$\mathcal{T}_k(x) = g \big[ \max(1, \norm{1}{x}-k) \big] $$ \\

Now that we obtained a close formula for $\mathcal{T}_k(x)$, we can determine : 
\[  \smooth_{ \mathcal{G}, \beta}(x)  = \max_{k \in \mathbb{N}} \ e^{-k \beta} \mathcal{T}_k(x) 
 =\max_{k \in \mathbb{N}} \ e^{-k \beta} \cdot g \big[ \max(1, \norm{1}{x}-k) \big]   \]

Let \[\xi_{x,\beta}(t): \left|   \begin{array}{ccl}
    \mathbb{R}^+ & \longrightarrow  & \mathbb{R}^+\\
    t & \longmapsto & e^{-t \beta} \cdot g \big[ \max(1, \norm{1}{x}-t) \big]  \\
  \end{array} \right. \]

\[   
\xi_{x,\beta}(t) = 
\begin{cases}
e^{-t \beta} \cdot g(1) &\quad\text{if } t \geq \norm{1}{x} -1 \\
e^{-t \beta} \cdot g(\norm{1}{x}-t)   &\quad\text{if } t \leq \norm{1}{x} -1 \\
\end{cases}
\]

$\xi_{x,\beta}$ is continuous on $\mathbb{R}^+$ and differentiable on $[0,\norm{1}{x} -1[$ and $]\norm{1}{x} -1, +\infty[$. 
The monotonicity of $\xi_{x,\beta}$ is trivial for high values of $t$:
   \[ \forall t \in ]\norm{1}{x} -1, +\infty[, \xi_{x,\beta}'(t) = - \beta \times e^{-t\beta} g(1) < 0\]

 $\forall t \in [0, \norm{1}{x} -1[,$

\[\begin{WithArrows}
 \xi_{x,\beta}'(t) &= - \beta \times e^{-t\beta} g(\norm{1}{x} -t) + e^{-t\beta} \times (-1) \times g'(\norm{1}{x} -t)   \Arrow{$y \coloneqq \norm{1}{x} -t $} \\
 &=  -  e^{-t\beta} \big[ \beta g(y) + g'(y) \big] \\
 &= -  e^{-t\beta} \Big[ \beta \Big(1 - \dfrac{y^2}{(y+1)^2} - \dfrac{1}{(y+1)^2} \Big)  + \dfrac{2(1-y)}{(y+1)^3} \Big] \\
 &= -  e^{-t\beta} \times \dfrac{\beta \cdot(y+1)^3 - \beta \cdot y^2 (y+1) - \beta \cdot (y+1) + 2(1-y)} {(y+1)^3}  \\
 &= \dfrac{ e^{-t\beta}}{(1+y)^3} \times \Big[-\beta \cdot(y+1)^3 + \beta \cdot y^2 (y+1) + \beta \cdot (y+1) - 2(1-y) \Big]
\end{WithArrows} \]

Since $\dfrac{ e^{-t\beta}}{(1+y)^3} > 0$ the sign of $\xi_{x,\beta}'(t)$ on $ [0, \norm{1}{x} -1[$ is given by the polynomial $P(Y) = -\beta \cdot(Y+1)^3 + \beta \cdot Y^2 (Y+1) + \beta \cdot (Y+1) - 2(1-Y)= -2 \beta Y^2 + (2-2\beta)Y - 2$. \newline

Let $Q \coloneqq - \beta Y^2 + (1-\beta)Y - 1 = P/2$. $P$ and $Q$ share the same roots, we will therefore study $Q$.
Let $\Delta$ the discriminant of polynomial $Q$. 
We associate it to the function $\Delta(\beta)$ since its value depends on $\beta$. 
The value of the discriminant gives whether or not the underlying function is monotonous.
$\Delta(\beta) = (1-\beta)^2 - 4 \beta = (\beta - 3 - 2 \sqrt{2}) (\beta - 3 + 2 \sqrt{2})$\newline

%\begin{adjustbox}{width=\linewidth} 
\begin{tikzpicture}
   \tkzTabInit{$\beta$ / 1 , $\Delta(\beta)$ / 1}{$0$, $ \beta_1 \coloneqq 3 - 2 \sqrt{2}$,$ \beta_2 \coloneqq 3 + 2 \sqrt{2}$,$+\infty$}
   \tkzTabLine{ ,+,z, -,z,+, }
\end{tikzpicture}
%\end{adjustbox}

%\vspace{2cm}

\begin{itemize}
    \item For $\beta \in ]3 - 2 \sqrt{2}, 3 + 2 \sqrt{2}[$, $\Delta(\beta) <0$ so $Q$ has no roots in $\mathbb{R}$ so it is negative on $\mathbb{R}$.
    \begin{tikzpicture}
   \tkzTabInit{$t$ / 1 , $\xi_{x,\beta}'(t)$ / 1, $\xi_{x,\beta}$ /2}{$0$, $\norm{1}{x}-1$,$+\infty$}
   \tkzTabLine{ ,-, -, -, }
   \tkzTabVar{+/$g(\norm{1}{x})$, R/, -/ $0$}
   \tkzTabIma{1}{3}{2}{$\frac{\exp(-(\norm{1}{x}-1)\beta)}{2}$}
   \end{tikzpicture}

    In that scenario, \fbox{$S_{ \mathcal{G}, \beta}(x) =  \xi_{x,\beta}(0) = g(\norm{1}{x})= LS_\mathcal{G}(x)$}
    
    \item For $\beta =3 - 2 \sqrt{2}$ or $\beta =3 + 2 \sqrt{2}$, $\Delta(\beta) = 0$ so $Q$ admits a unique root $y_0 = \frac{1-\beta}{2\beta}$ (we will ignore these two values of $\beta$ as there are enough $\beta$ that we can choose).
    
    \item For $\beta \in ]0, 3 - 2 \sqrt{2}[ \cup ]3+ 2 \sqrt{2}, + \infty[$, $\Delta(\beta) > 0$ so $Q$ admits two distinct roots : 
    \[ y_1 = \dfrac{1-\beta +  \sqrt{(1-\beta)^2 - 4 \beta} } {2 \beta} \qquad \text{and}  \qquad y_2 = \dfrac{1-\beta -  \sqrt{(1-\beta)^2 - 4 \beta} } {2 \beta} \]

%\begin{adjustbox}{width=\linewidth}
\begin{tikzpicture}
   \tkzTabInit{$y$ /1, $Q(y)$ / 1}{ -$\infty$,$y_2$, $y_1$,$+\infty$}
   \tkzTabLine{ ,-,z,+,z,-,}
\end{tikzpicture}
%\end{adjustbox}

  The problem is that the roots $t_1 \coloneqq \norm{1}{x} - y_1$ and $t_2 \coloneqq \norm{1}{x} - y_2$ might overflow the interval $ [0, \norm{1}{x} -1[$. Since $y \mapsto \norm{1}{x} - y \eqqcolon t $ is a strictly decreasing function (it is a bijection from $\mathbb{R}$ to $\mathbb{R}$) we have that $y_2< y_1 \implies t_2 > t_1$. 
  What we want to study is the mapping  from $[y_2, y_1]$ to $[t_1, t_2]$ with respect to the domain of validity for the studied form of $\xi_{x,\beta}$. 
  A case per case analysis (detailed below) shows that the smooth sensitivity of the Gini for these values of $\beta$ is given by the formula: 

\[
\boxed{
  \!\begin{aligned}
  \smooth_{ \mathcal{G}, \beta}(x) &=  \max\Big[\xi_{x,\beta}(0), \xi_{x,\beta}(\floor{t_2}),\xi_{x,\beta}(\ceil{t_2})\Big] \\
 &=  \max\Big[g(\norm{1}{x}), e^{-\floor{t_2} \beta} g(\norm{1}{x}-\floor{t_2}),e^{-\ceil{t_2} \beta} g(\norm{1}{x}-\ceil{t_2})\Big]
  \end{aligned}
}
\] \\

We first compute the values of the roots $y_1$ and $y_2$ according to $\beta$ (in particular the asymptotic values).

$$ y_1 \equivasym{\beta \to 0} \dfrac{1}{\beta} \xrightarrow[\beta \to 0]{} +\infty \qquad \text{and} \qquad y_1 \equivasym{\beta \to +\infty} \dfrac{1}{1-\beta} \xrightarrow[\beta \to +\infty]{} 0^- $$

      $$  y_2 \equivasym{\beta \to 0} \dfrac{1}{(1-\beta)^2} \xrightarrow[\beta \to 0]{}  1 \qquad \text{and} \qquad y_2 \equivasym{\beta \to +\infty} -1  \qquad \quad$$

 %\begin{adjustbox}{width=\linewidth}
\begin{tikzpicture}
   \tkzTabInit{$\beta$ /1, $y_1(\beta)$ / 2, $y_2(\beta)$/2}{$0$, $3 - 2 \sqrt{2}$, $3 + 2 \sqrt{2}$, $+\infty$}
   \tkzTabVar{+/ $+\infty$, -DH/ $3> \cdot >2$, D-/$-1<\cdot<0$, +/ $1^-$}
   \tkzTabVar{-/ $1$, +DH/ $
   3> \cdot >2$, D+/$-1<\cdot<0$, -/ $-1$}
\end{tikzpicture}
%\end{adjustbox}
That gives us two cases to treat:

If $\beta \in ]0, \beta_1[ $. 
In the case that $\norm{1}{x} \geq 5$ (which is a reasonable assumption)  $\exists \beta^* \in ]0, \beta_1[, \forall \beta \geq \beta^*, 0< t_1(\beta) < \norm{1}{x} -1 $ and  $0<t_2 < \norm{1}{x} - 1 $ (for all $\beta$ in the considered interval) which gives : $0 < t_1 < t_2 < \norm{1}{x} -1$
    \begin{itemize}
        \item So if $\beta$ is too small, then the $t$'s associated to the interval $[y_2, y_1]$ are ( $<0$) partly outside the domain of validity which yields 

 %\begin{adjustbox}{width=\linewidth}
\begin{tikzpicture}
   \tkzTabInit{$t$ /1, $Q(t)$ / 1, $\xi_{x,\beta}$/2}{$t_1$, 0, $t_2$, $\norm{1}{x} -1$ ,$+\infty$}
   \tkzTabLine{ z,,+,,z,-,d,h,}
\tkzTabVar{-DH/,-/$g(\norm{1}{x})$, +/$\xi_{x,\beta}(t_2)$, R/,-/0}
\end{tikzpicture} 
%\end{adjustbox}\\

Hence :
\[
\boxed{
  \!\begin{aligned}
  \smooth_{ \mathcal{G}, \beta}(x) &=  \max\Big[\xi_{x,\beta}(\floor{t_2}),\xi_{x,\beta}(\ceil{t_2})\Big] \\
  \smooth_{ \mathcal{G}, \beta}(x) &=  \max\Big[e^{-\floor{t_2} \beta} g(\norm{1}{x}-\floor{t_2}),e^{-\ceil{t_2} \beta} g(\norm{1}{x}-\ceil{t_2})\Big]
  \end{aligned}
}
\]

   \item if $ \beta \in ]\beta^*,\beta_1[$, then all the $t$'s associated to $[y_2, y_1]$ are in the domain of validity.

% \begin{adjustbox}{width=\linewidth}
\begin{tikzpicture}
   \tkzTabInit{$t$ /1, $Q(t)$ / 1, $\xi_{x,\beta}$/2}{ 0,$t_1$, $t_2$, $\norm{1}{x} -1$ ,$+\infty$}
   \tkzTabLine{ ,-,z,+,z,-,d,h}
\tkzTabVar{+/$g(\norm{1}{x})$, -/$\xi_{x,\beta}(t_1)$ , +/$\xi_{x,\beta}(t_2)$, R/,-/0}
\end{tikzpicture} 
%\end{adjustbox}

\[
\boxed{
  \!\begin{aligned}
  \smooth_{ \mathcal{G}, \beta}(x) &=  \max\Big[\xi_{x,\beta}(0), \xi_{x,\beta}(\floor{t_2}),\xi_{x,\beta}(\ceil{t_2})\Big] \\
  \smooth_{ \mathcal{G}, \beta}(x) &=  \max\Big[g(\norm{1}{x}), e^{-\floor{t_2} \beta} g(\norm{1}{x}-\floor{t_2}),e^{-\ceil{t_2} \beta} g(\norm{1}{x}-\ceil{t_2})\Big]
  \end{aligned}
}
\]

    \end{itemize}

If $\beta \in ]\beta_2, + \infty[$.   $t_2 > \norm{1}{x} - 1 $ and $ t_1> \norm{1}{x} -1 $ which means that the $t$'s associated to the $[y_2, y_1]$ are ( $> \norm{1}{x}-1$) all outside the domain of validity. 

%\begin{adjustbox}{width=\linewidth}
\begin{tikzpicture}
   \tkzTabInit{$t$ /1, $Q(t)$ / 1, $\xi_{x,\beta}$/2}{ 0, $\norm{1}{x} -1$ ,$t_1$, $t_2$,$+\infty$}
   \tkzTabLine{ ,-,d,h,d,h,d,h,}
\tkzTabVar{+/$g(\norm{1}{x})$,R/ ,R/,R/,-/0}
\end{tikzpicture} 
%\end{adjustbox}\\

\fbox{$\smooth_{ \mathcal{G}, \beta}(x) =  \xi_{x,\beta}(0) = g(\norm{1}{x})= LS_\mathcal{G}(x)$}
\end{itemize}

%\vspace{0.5cm}

\subsection{Case 2: Generalization. Assume now that $\Lambda \in \mathbb{N}^*$}
\label{appendix:smooth-sensitivity-lambda}

It is easy to prove that: $\mathcal{T}_k(x) = g \big[ \max(\Lambda, \norm{1}{x}-k) \big]$. Therefore, we define the function:
\[\xi_{x,\beta}(t): \left|   \begin{array}{ccl}
    \mathbb{R}^+ & \longrightarrow  & \mathbb{R}^+\\
    t & \longmapsto & e^{-t \beta} \cdot g \big[ \max(\Lambda, \norm{1}{x}-t) \big]  \\
  \end{array} \right. \]

\[   
\xi_{x,\beta}(t) = 
\begin{cases}
e^{-t \beta} \cdot g(\Lambda) &\quad\text{if } t \geq \norm{1}{x} - \Lambda \\
e^{-t \beta} \cdot g(\norm{1}{x}-t)   &\quad\text{if } t \leq \norm{1}{x} - \Lambda \\
\end{cases}
\]

$\xi_{x,\beta}$ is continuous on $\mathbb{R}^+$ and differentiable on $[0,\norm{1}{x} -\Lambda[$ and $]\norm{1}{x} -\Lambda, +\infty[$. $\xi_{x,\beta}$ derivatives remain unchanged but the bounds are shifted (from 1 to $\Lambda$). The roots $y_1$ and $y_2$ are unchanged (they solely depend on $\beta$). Instead of re-doing the case per case analysis, we will propose the following exact method.

\begin{enumerate}
    \item Compute $t_1$ and $t_2$ if the roots $y_1$ and $y_2$ exist
    \item Compute the relative positions of $t_1$ and $t_2$ with respect to $0$ and $\norm{1}{x} - \Lambda$
    \item We know that  $\xi_{x,\beta}$ is increasing between $t_1$ and $t_2$ granted that they are in the $[0,\norm{1}{x} - \Lambda[$ interval so there is an eventual max in this interval, to compare to $\xi_{x,\beta}(0)$ and  $\xi_{x,\beta}(\norm{1}{x} - \Lambda)$ 
\end{enumerate}
This achieves the proof of Theorem~\ref{theorem-smooth}. \newline
%\vspace{1em}
%\newpage

%\newpage
%\section{Additional Results}

%\subsection{Additional Figures for Section~\ref{subsec:expe-global-gini}}
%\label{subsec:more-figures-global}

\section{Membership Inference Attacks - Detailed Results}
\label{subsec:mia-results}  

\begin{figure*}[h!]
    \centering
    \includegraphics[scale=0.4]{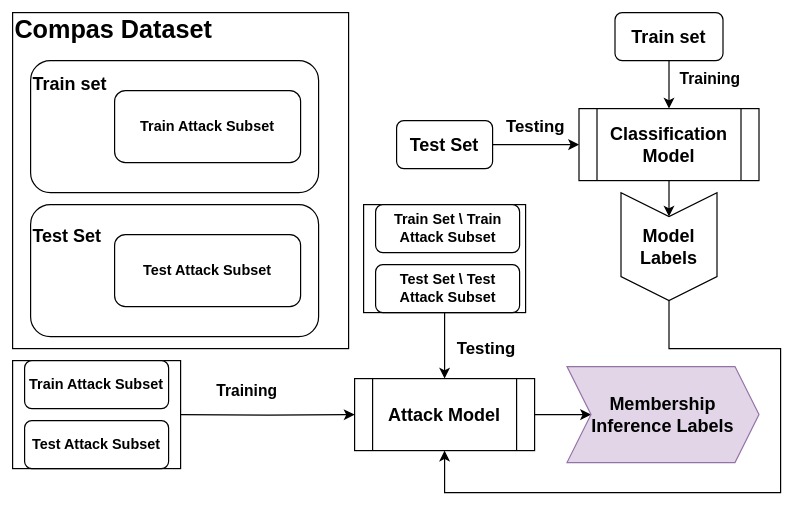}
    \caption{Pipeline of Membership Inference Attack}
    \label{fig:mia-pipeline}
\end{figure*}

Figure~\ref{fig:mia-pipeline} illustrates how a dataset is split into different subsets to train a Membership Inference Attack (MIA) model. 
Note that the $\setminus$ symbol represents the minus operation on sets (set difference).

\begin{figure*}[b!]
  \centering
  \begin{subfigure}[t]{0.48\textwidth}
    \centering\includegraphics[width=\textwidth]{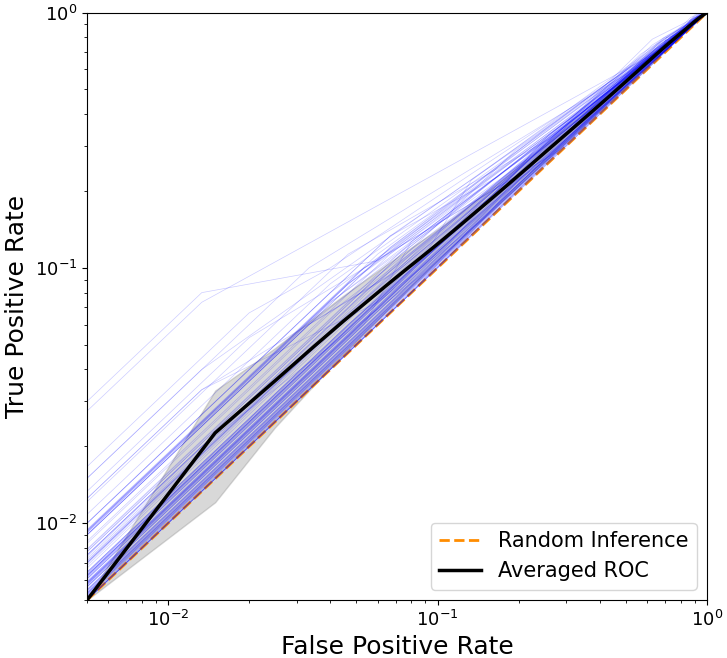}
    \caption{\namedataset{\namedataset{German credit} - \GRL{} }}
  \end{subfigure}
  \hfill
  \begin{subfigure}[t]{0.48\textwidth}
    \centering\includegraphics[width=\textwidth]{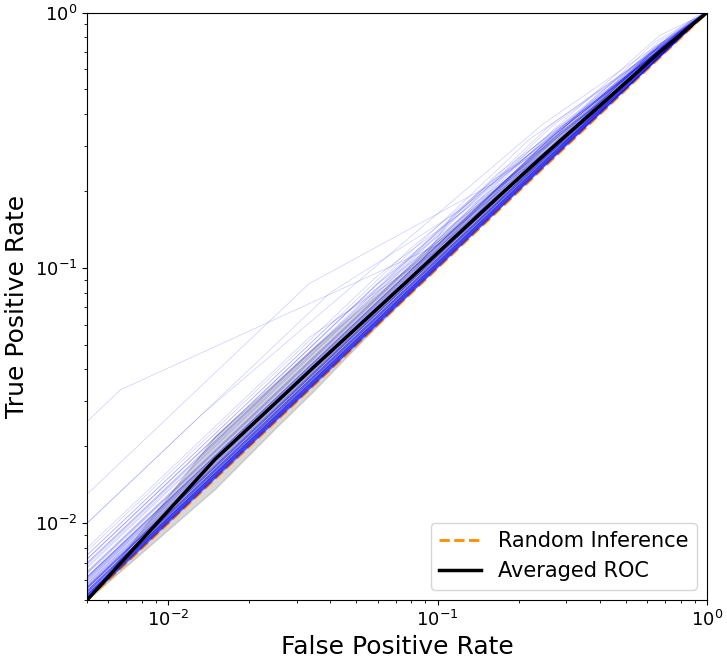}
    \caption{\namedataset{\namedataset{German credit} - \DPGRL}}
  \end{subfigure}
  \vspace{1em}
  \begin{subfigure}[t]{0.48\textwidth}
    \centering\includegraphics[width=\textwidth]{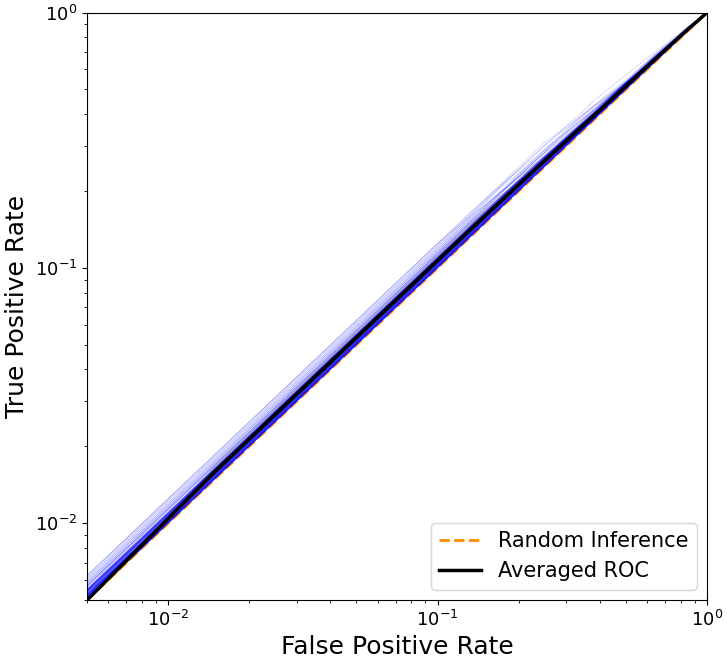}
    \caption{\namedataset{\namedataset{Compas} - \GRL{}}}
  \end{subfigure}  
  \hfill
  \begin{subfigure}[t]{0.48\textwidth}
    \centering\includegraphics[width=\textwidth]{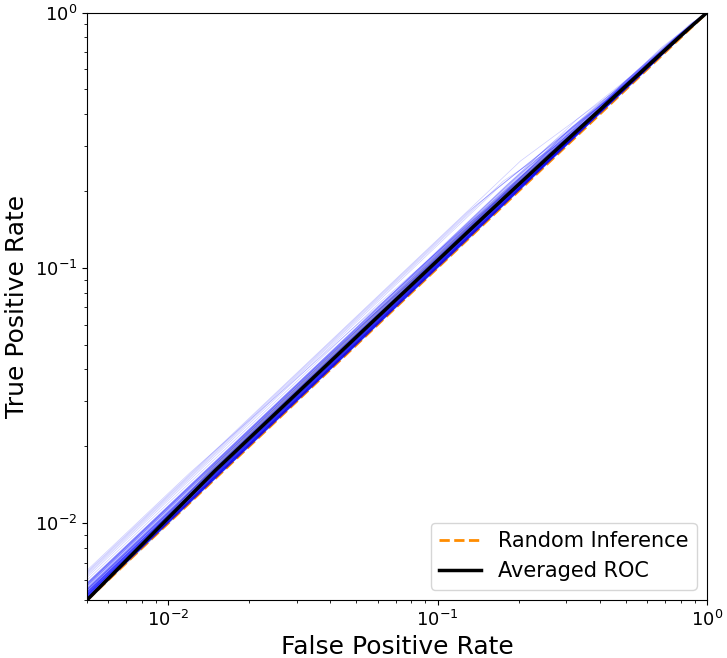}
    \caption{\namedataset{\namedataset{Compas} - \DPGRL}}
  \end{subfigure}
  \caption{ROC Curves of Membership Inference Attacks on the DP \DPGRL{} and on the baseline \GRL{} models.}
\label{fig:roc}

\end{figure*}

We consider two MIAs from the ART toolkit\footnote{\url{https://github.com/Trusted-AI/adversarial-robustness-toolbox/wiki/ART-Attacks\#4-inference-attacks}}~\cite{art}. %\jul{Voir si ils demandent une citation}
The results we obtained for a black-box MIA\footnote{\url{https://adversarial-robustness-toolbox.readthedocs.io/en/latest/modules/attacks/inference/membership_inference.html\#membership-inference-black-box}} using Random Forests are presented in Figure~\ref{fig:roc}. 
The ROC curves are displayed in log scale to highlight the results at low FPR since it is the relevant regime for MIAs~\cite{carlini_mia}. 
They show that, as mentioned in Section~\ref{subsec:expe-attacks}, (even non-DP) rule lists are already resilient to MIAs. 
On the smallest \namedataset{German credit} dataset, we observed a slightly higher distributional overfitting, which results in slightly higher TPRs at low FPR. 

%\ctim{quelques mots sur les attaques d'inférence, fonctionnement sans rentrer dans les détails techniques}

%\newpage
We also considered the Label Only Membership Inference Attack\footnote{\url{https://adversarial-robustness-toolbox.readthedocs.io/en/latest/modules/attacks/inference/membership_inference.html\#membership-inference-label-only-decision-boundary}}~\cite{choquetteabelonly} but results were sub-par due to the datasets used. 
Indeed, the rule lists use as input binarized features whereas the attack explores the latent variables space by studying how the model output varies when the features values are tweaked. 
The issue here is that the model can only read features that are $0$ or $1$ and therefore we had to truncate the latent space exploration to the much sparser space of $\{0,1\}^m$, making it inefficient. 
In addition, since the datasets are binarized, some features are actually a one-hot-encoding of a categorical feature, which means it does not make sense that several of them can be set to 1. 
An interesting avenue of research would be to use the latent space exploration on the non-binarized features and re-apply the binarization process at each step. 
This is unfortunately computationally expensive and would likely lead to similar results due to the loss of information incurred by the binarization step preceding inference by the model.

\end{document}